%% file: main.tex
\pdfoutput=1

\documentclass[conference]{IEEEtran}

\def\compileFigures{0}
\newcommand{\filename}{main}
\newcounter{figureNumber}

\if\compileFigures1
\usetikzlibrary{external}
\usetikzlibrary{calc}
\usetikzlibrary{positioning}
\fi

\input{packages_and_commands.tex}

\input{macros}

\title{Range Membership Inference Attacks}

\author{\IEEEauthorblockN{Jiashu Tao}
	\IEEEauthorblockA{
		\textit{National University of Singapore}\\
		jiashut@comp.nus.edu.sg}
	\and
	\IEEEauthorblockN{Reza Shokri}
	\IEEEauthorblockA{
		\textit{National University of Singapore}\\
		reza@comp.nus.edu.sg}
}

\begin{document}
\maketitle

\input{sections/abstract.tex}
\input{sections/sec_intro.tex}
\input{sections/sec_preliminary.tex}
\input{sections/sec_formulation.tex}
\input{sections/sec_range_mia.tex}
\input{sections/sec_experiments.tex}
\input{sections/sec_conclusion.tex}
\input{sections/acknowledgement.tex}

\bibliographystyle{IEEEtran}
\bibliography{IEEEabrv, reference}

\appendix
\input{sections/app_attacks.tex}
\input{sections/app_more_results.tex}

\end{document}

%% file: packages_and_commands.tex
\usepackage{cite}
\usepackage{amsmath}
\usepackage{amssymb}
\usepackage{mathtools}
\usepackage{hyperref}       
\usepackage{url}            
\usepackage{booktabs}       
\usepackage{amsfonts}       
\usepackage{nicefrac}       
\usepackage{microtype}      
\usepackage{xcolor}         

\usepackage{bm}

\usepackage{graphicx}
\usepackage{caption}
\usepackage{subcaption}
\usepackage{comment}
\usepackage{multirow}
\usepackage{algorithm}
\usepackage{algpseudocode}
\usepackage{enumitem}

\usepackage[colorinlistoftodos]{todonotes}
\usepackage{tikz}
\usepackage{pgfplots}
\usepgfplotslibrary{fillbetween}
\usepackage{rotating}

\definecolor{matplotblue}{RGB}{31,119,180}
\definecolor{matplotorange}{RGB}{255,127,14}
\definecolor{matplotgreen}{RGB}{44,160,44}
\definecolor{matplotred}{RGB}{214,39,40}
\definecolor{matplotpurple}{RGB}{148,103,189}
\definecolor{matplotbrown}{RGB}{140,86,75}
\definecolor{matplotpink}{RGB}{227,119,194}
\definecolor{matplotgrey}{RGB}{127,127,127}
\definecolor{matplotbgreen}{RGB}{188,189,34}
\definecolor{matplotcyan}{RGB}{23,190,207}

\definecolor{col1}{RGB}{237,248,177}
\definecolor{col2}{RGB}{127,205,187}
\definecolor{col3}{RGB}{44,127,184}

%% file: macros.tex
\newcommand{\x}[0]{\mathbf{x}}

\newcommand{\probP}{\text{I\kern-0.15em P}}

\usepackage{mathtools}

\DeclarePairedDelimiterX{\infdivx}[2]{(}{)}{%
  #1\;\delimsize\|\;#2%
}

\newtheorem{definition}{Definition}

%% file: sections/abstract.tex
\begin{abstract}
Machine learning models can leak private information about their training data. The standard methods to measure this privacy risk, based on membership inference attacks (MIAs), only check if a given data point \textit{exactly} matches a training point, neglecting the potential of similar or partially overlapping memorized data revealing the same private information. To address this issue, we introduce the class of range membership inference attacks (RaMIAs), testing if the model was trained on any data in a specified range (defined based on the semantics of privacy). We formulate the RaMIAs game and design a principled statistical test for its composite hypotheses. We show that RaMIAs can capture privacy loss more accurately and comprehensively than MIAs on various types of data, such as tabular, image, and language. RaMIA paves the way for more comprehensive and meaningful privacy auditing of machine learning algorithms. 

\end{abstract}


%% file: sections/sec_intro.tex
\section{Introduction}
Machine learning models are prone to training data memorization \cite{feldman2019does,feldman2020neural, liu2021understanding, tirumala2022memorization, kim2023memorization}. It is also a known fact that the outstanding predictive performance of machine learning models on long-tailed data distributions often comes at the expense of blatant memorization of certain data points \cite{feldman2020neural, brown2021memorization, lukasik2023larger, garg2023memorization}. Memorization refers to the phenomenon where models behave differently on data depending on whether it was included in the training set. Such behavior can lead to significant privacy risks because adversaries can infer sensitive information about training data even with only black-box access to the model.

To quantify the privacy risk of machine learning models, it is crucial to define a precise privacy notion. The prevailing privacy notion is based on \textit{membership} information, a binary indicator that carries substantial privacy implications. Accurate inference of membership status can enable data \textit{reconstruction} attacks~\cite{salem2020updates, hilprecht2019monte, carlini2021extracting, long2023membership}, where the adversary probes the membership of plausible data points to recover the training set. The de facto way to audit the privacy risk according to this privacy notion is to conduct membership inference attacks (MIAs)~\cite{shokri2017membership}, where an adversary aims to predict whether a given point belongs to the training set of the target model. The more powerful the membership inference attack is, the higher the privacy risk the target model bears. 

\begin{figure*}[!bt]
	\centering
	\includegraphics[width=\textwidth]{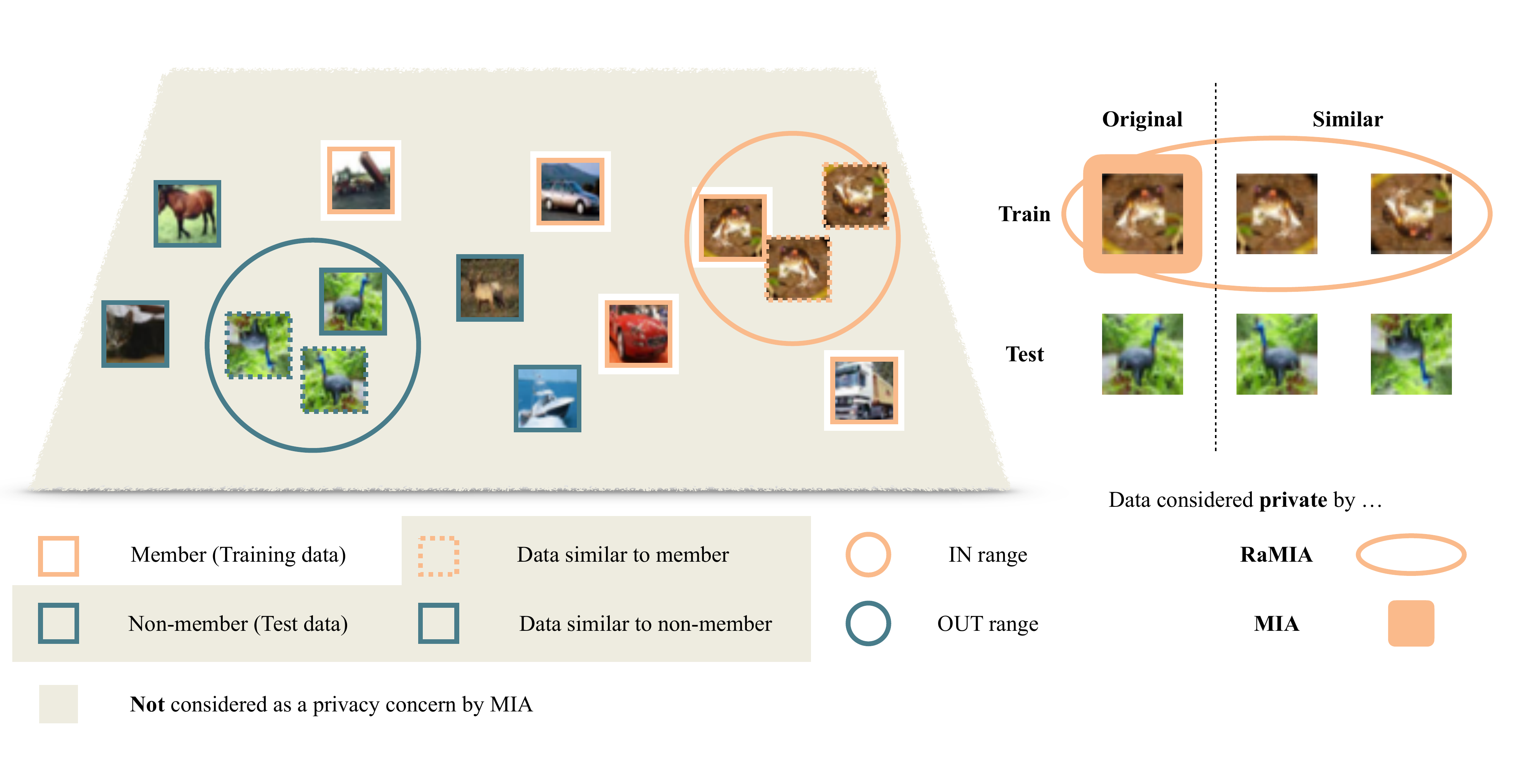}
	\caption{Illustration of privacy notions for MIAs and RaMIAs. \textbf{LEFT:} Sample CIFAR-10 training and test data are depicted with one highlighted IN range and one OUT range for visualization. In MIAs, only exact training points are deemed private, while all other points in the data space are considered non-private. The privacy notion is hence akin to a Dirac delta function that assigns privacy only to exact training points and zero everywhere else. \textbf{RIGHT:} RaMIA expands this definition by treating points that are similar to training data as privacy leaking. This approach captures privacy risks not only at exact training points but also in their vicinity, offering a more comprehensive and realistic privacy assessment.} 
	\label{fig:ramia_vs_mia}
\end{figure*}

\begin{figure*}[!htb]
	\centering
		\begin{subfigure}[b]{0.3\textwidth}
				\centering
				\includegraphics[width=0.9\textwidth]{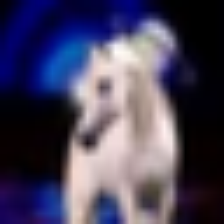}
				\caption{MIAScore=2.99}
				\label{fig:cifa_example}
			\end{subfigure}
		\hfill
		\begin{subfigure}[b]{0.3\textwidth}
				\centering
				\includegraphics[width=0.9\textwidth]{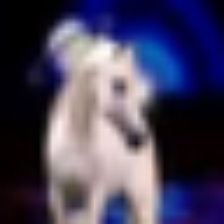}
				\caption{MIAScore=2e-15}
				\label{fig:cifa_example_flipped}
			\end{subfigure}
		\hfill
		\begin{subfigure}[b]{0.38\textwidth}
				\centering
				\if\compileFigures1
				\input{figure_scripts/text_examples.tex}
				\else
				\includegraphics[]{fig/\filename-figure\thefigureNumber.pdf}
				\stepcounter{figureNumber}
				\fi
				\caption{MIAScore=1.8 and 1.0}
				\label{fig:text_example}
			\end{subfigure}
	\caption{Examples of similar data with drastically different MIA scores and potentially opposite membership predictions. The dog image is from the CIFAR-10 dataset, while the text data is from AG News. The image classifier is not trained with horizontal flipping augmentation.}
	\label{fig:example}
\end{figure*}

Membership inference attacks provide a \textit{lower bound} of the model's true privacy risk, so improving the attack performance also tightens the bound of privacy risk estimation. So far, to more accurately audit the privacy risk, the community has largely focused on enhancing MIAs by developing stronger membership signals and more sophisticated statistical tests \cite{sankararaman2009genomic, shokri2017membership, sablayrolles2019white, ye2022enhanced, carlini2022membership, zarifzadeh2023low}. While these advances have improved privacy auditing, they all quantify privacy risks by testing memorization of the \textbf{exact, full version} of training points. However, information leakage is not an all or nothing phenomenon. In many cases, a model may not memorize a training record in its entirety but rather partial information in the form of key features or data patterns. These memorization behaviors, which are more realistic and prevalent, lead to a form of leakage of sensitive information that can be exhibited from similar points in the vicinity of training data. For example, a model trained on images including people's faces may capture the distinctive facial attributes that uniquely identify an individual while ignoring extraneous details such as the background. This could be a concern, as an image of the same person taken from a different angle or with a different background has a significant overlap in private information. Membership inference attacks, designed to detect exact matches, are not equipped to quantify this notion of information leakage as any slight change to training points would turn them into non-members where any correct MIA is expected to produce negative outcomes. For example, when using MIA, a simple horizontal flip can reduce a membership score from a high value to zero (Figure \ref{fig:example}), and overall AUC can drop by 20\% when testing image classifiers with horizontally flipped images (Figure \ref{fig:mia_cifar}). This is expected as the transformed images are, by definition of MIA, non-members. 

To address this gap, we propose a new class of inference attack, \emph{\textbf{Ra}nge \textbf{M}embership \textbf{I}nference \textbf{A}ttack} (\textbf{RaMIA}), which is specifically designed to audit this new type of information leakage (Fig \ref{fig:ramia_vs_mia}). The goal of RaMIA is to \textit{determine whether any training data exists within a defined neighborhood around a candidate point}. Instead of relying on point queries that seek an exact match, RaMIA uses range queries that are defined by a center point, a distance measure that captures the semantics of privacy, and a radius that describes the area of interest. In practice, an auditor can craft a range query centered on any data record, image, or text. By applying a distance function that preserves the sensitive features (e.g., using the $\ell_2$ distance on unimportant features or tokens), the auditor can test whether the model has memorized these sensitive features even when the data tested are not exact replicas of any training data. In this way, privacy auditors can more accurately and comprehensively assess the privacy risk associated with sensitive information by tailoring the center, range function, and radius to address their specific concerns.

RaMIA is a flexible framework that supports a wide range of distance measures, enabling it to quantify various privacy risks based on different notions of proximity. Moreover, RaMIA aligns more closely with our intuitive understanding of privacy risk: when assessing the privacy of a data point, we expect the test to go beyond detecting exact matches and instead capture if any sensitive information is being memorized. For instance, if we claim that a model does not leak private information about my photo, we expect that it does not reveal any private information about me instead of merely the exact test photo. The range MIA enables formulating this notion of privacy risk. 

RaMIA is also naturally related to current evaluation protocols for data extraction attacks on generative models \cite{carlini2021extracting, wu2022membership, carlini2023extracting}, where candidate data points that fall within a specific tolerance are treated as successful extractions, reflecting the fact that close-by points can expose private information. Moreover, RaMIA can bypass naive test-time privacy protection schemes in which inputs that are recognized as training data are filtered (to fail naive use of MIA). Since the model disrupts the correspondence between inputs and exact training points, MIA is not expected to work properly. On the other hand, because RaMIA is capable of detecting information leakage in points close to training data, it can easily bypass this simple defense, thereby offering a more robust way of auditing information leakage against data reconstruction attacks.

Range membership inference attacks extend the formulation of exact membership inference attacks by incorporating range queries that capture privacy leakage in the vicinity of training data. We modify the traditional membership inference game to accommodate range queries, leading to composite hypotheses in the likelihood ratio tests, which are the standard and most effective techniques in MIAs \cite{sankararaman2009genomic, ye2022enhanced, carlini2022membership, zarifzadeh2023low}. Our method leverages robust statistical approaches for composite hypothesis testing, including generalized likelihood ratio tests (GLRTs) and Bayes factors. We show that RaMIA provides a more comprehensive notion of privacy by detecting leakage from nearby training data when the direct application of MIAs underestimates the true risk. For example, testing horizontally flipped images instead of the original training and test images leads to a 20\% drop in AUC when audited by MIAs (see Fig \ref{fig:mia_cifar}). As a proof-of-concept, we implement RaMIA with a straightforward attack strategy and conduct experiments on tabular, image, and text datasets, where RaMIA consistently outperforms traditional MIA. Notably, even in the most challenging scenario where each IN range contains only one training point positioned near the boundary, our simple RaMIA algorithm (Sec \ref{sec:ramia}) achieves a 5\% improvement on image datasets (see Figs \ref{fig:attack_celeba} and \ref{fig:attack_cifar}). This result is obtained using at most 15 samples, which represents a negligible cost given the high dimensionality of the data space. It is important to note that these gains are observed when comparing RaMIA with the case where only traditional MIA is available and the MIA score of the center point is used to audit privacy with RaMIA's membership definition. RaMIA’s performance further improves in less restrictive settings where each IN range contains multiple training points, underscoring its robustness and practical impact for comprehensive privacy auditing.

%% file: figure_scripts/text_examples.tex
\begin{tikzpicture}
	\centering
	\node[draw, rounded corners, text width=5cm] (sentence1) {Judging Error Led to Hamm's Gold ATHENS, Greece - Paul Hamm thought his fantastic finish was too good to be true. Maybe he was \textcolor{red}{right...}};

	\node[draw, rounded corners, below=0.2cmof sentence1, text width=5cm] (sentence2) {Judging Error Led to Hamm's Gold ATHENS, Greece - Paul Hamm thought his fantastic finish was too good to be true. Maybe he was \textcolor{red}{?}};
\end{tikzpicture}

%% file: sections/sec_preliminary.tex
\section{Related Work}
\subsection{Membership inference attacks} \label{sec:mia}
The membership inference attack (MIA) \cite{shokri2017membership} is a class of inference attack against machine learning models to infer whether a given data sample is part of the model's training set. Mathematically, given a model $f$ and a query point $\x$, the MIA aims to output 1 if $\x$ is a training point, and 0 otherwise. Over time, various methods have been developed to construct and execute MIAs, making it an active research area with continuously evolving techniques. Shokri et al.~\cite{shokri2017membership} introduced a shadow model-based approach. In this method, multiple shadow models are trained on datasets that mimic the target model’s training set, and the confidence values for both training and test data are computed to serve as benchmarks. However, the high cost and strong assumption of knowing the target model's training details make the attack often infeasible. \cite{yeom2018privacy} use model loss as a signal and threshold it, scraping the need for shadow models. Then MIA is formulated as an inference game (See Sec \ref{sec:mia_game}). Subsequent research has adopted a more principled approach by solving this inference game using likelihood ratio tests \cite{sankararaman2009genomic, carlini2022membership, ye2022enhanced, zarifzadeh2023low}. \cite{carlini2022membership} and \cite{ye2022enhanced} propose reference model-based approaches, where target signals are compared to those obtained on reference models to obtain the likelihood ratio. To further boost the attack power, \cite{zarifzadeh2023low} assumes the attacker has access to a pool of population data so that the likelihood ratio from reference-based attacks can be further calibrated on population data.

\paragraph{Membership inference attacks with augmentations} Recent attacks \cite{carlini2022membership, zarifzadeh2023low} find extra attack performance on image data from augmenting the test queries with \emph{train-time }augmentations, as these augmented training images could have been seen by the model during training. This approach, however, presumes that the attacker knows the exact train-time augmentations and can replicate them. Augmenting training images with non train-time augmentations is deliberately excluded, as such images would be classified as non-members under the current privacy notion, and testing them as members would contradict the established definition.

\paragraph{Perturbation-based membership inference attacks} Several approaches have explored utilizing neighboring points around the query to improve MIA performance. \cite{choquette2021label} hypothesize that membership scores should remain consistently high in the vicinity of training points, and they aggregate scores from neighboring points to assess the smoothness of the scoring function. In contrast, \cite{mattern2023membership} argue that the loss curvature would be sharper around training points, causing MIA scores to decline rapidly as one moves away from the training data. Although these perturbation-based methods incorporate neighboring information, they ultimately classify only the exact training points as members. Any perturbed or neighboring point is treated as non-member.

\subsection{Range queries} \label{sec:range_query}
Drawing a parallel with database systems, traditional MIAs operate on \emph{point queries} or \emph{exact match queries}, where a single data point is retrieved. In contrast, a range query is designed to retrieve all data points within a specified interval or “range.” The key difference is that a range query often returns multiple data points rather than a single exact match. Our proposed attack, the range membership inference attack (RaMIA), builds upon this concept by operating with range queries, thereby extending the conventional MIA framework to capture privacy leakage in a broader context.

%% file: sections/sec_formulation.tex
\section{From MIA to RaMIA}
Membership inference attacks (MIAs) have traditionally been formulated as an inference game~\cite{yeom2018privacy, jayaraman2021revisiting, ye2022enhanced, carlini2022membership, zarifzadeh2023low} between a \emph{challenger} and an \emph{adversary}. In this section, we first review the standard MIA framework and its evaluation protocol, then discuss its intrinsic limitations as a privacy auditing tool. These limitations motivate our proposed extension, \textbf{range membership inference attacks (RaMIAs)}, which broaden the notion of membership to include points that leak sensitive information while not exactly in the training set, .

\subsection{Membership inference attacks}
In membership inference attacks, the goal of the attacker is to identify if a given point is part of the training set.
\subsubsection{Membership inference game}  \label{sec:mia_game}
\begin{definition}(\textbf{Membership Inference Game } \cite{ye2022enhanced, yeom2018privacy}) \label{def:mia}
	Let~$\pi$ be the data distribution, and let $\mathcal{T}$ be the training algorithm.
	\begin{enumerate}
    	\item The \emph{challenger} samples a training dataset $D \longleftarrow \pi$, and trains a model $\theta \longleftarrow \mathcal{T}(D)$.
    	\item The \emph{challenger} samples a data record $z_0 \longleftarrow \pi$ from the data distribution, and a training data record $z_1  \longleftarrow D$.
    	\item The \emph{challenger} flips a fair coin to get the bit $b \in \{0, 1\}$, and sends the target model $\theta$ and data record $z_b$ to the \emph{adversary}.
    	\item The \emph{adversary} gets access to the data distribution $\pi$ and access to the target model, and outputs a bit $\hat{b} \longleftarrow \mathcal{A}(\theta, z_b)$.
    	\item If $\hat{b}=b$, output 1 (success). Otherwise, output 0.
	\end{enumerate}
\end{definition}

\subsubsection{Evaluation of MIA}
Conventionally, the MIA algorithm outputs a continuous MIA score for each point query. The membership decision is obtained by thresholding the score. Evaluation of MIAs is done on a set of training and test points. True positive rate (TPR) and false positive rate (FPR) are computed by sweeping over all  threshold values.  By plotting the receiver operating characteristic curve (ROC), the power of an attack strategy can be represented by the area under the curve (AUC). A clueless adversary who can only randomly guess the membership labels is expected to get an AUC of~$0.5$. Stronger adversaries predict membership more accurately at each error level. Hence, they would achieve higher TPR at each FPR, and get a higher AUC.

\subsubsection{Intrinsic limitation of MIA as a Privacy Auditing Framework} 
MIAs are \emph{designed} to be incapable of identifying points close to training points, regardless of how similar they are, due to the strict definition of members. Hence, there is a vast data space of points that contain private information but are deemed to be non-members in the current privacy auditing framework. In this way, MIAs as privacy auditing tools become out-of-scope when the queries move away from the original data, resulting in unpredictable and unreliable auditing results in these scenarios. Figure \ref{fig:mia_attack_performance} illustrates how MIA performance deteriorates as the query points deviate from the original training data. This observation motivates our formulation of RaMIA, where we broaden the definition of membership to capture privacy leakage from similar data points.

\begin{figure*}[!ht]
	\centering
	\begin{subfigure}[b]{0.49\textwidth}
		\centering
		\if\compileFigures1
		\input{figure_scripts/purchase_ramia_vs_mia.tex}
		\else
		\includegraphics[]{fig/\filename-figure\thefigureNumber.pdf}
		\stepcounter{figureNumber}
		\fi
		\caption{Purchase-100}
		\label{fig:mia_purchase}
	\end{subfigure}
	\hfill
	\begin{subfigure}[b]{0.49\textwidth}
		\centering
		\if\compileFigures1
		\input{figure_scripts/celeba_ramia_vs_mia.tex}
		\else
		\includegraphics[]{fig/\filename-figure\thefigureNumber.pdf}
		\stepcounter{figureNumber}
		\fi
		\caption{CelebA}
		\label{fig:mia_celeba}
	\end{subfigure}
	\vskip\baselineskip
	\begin{subfigure}[b]{0.49\textwidth}
		\centering
		\if\compileFigures1
		\input{figure_scripts/cifar10_matched_vs_unmatched_auc_clean_model.tex}
		\else
		\includegraphics[]{fig/\filename-figure\thefigureNumber.pdf}
		\stepcounter{figureNumber}
		\fi
		\caption{CIFAR-10}
		\label{fig:mia_cifar}
	\end{subfigure}
	\hfill
	\begin{subfigure}[b]{0.49\textwidth}
		\centering
		\if\compileFigures1
		\input{figure_scripts/agnews_ramia_vs_mia.tex}
		\else
		\includegraphics[]{fig/\filename-figure\thefigureNumber.pdf}
		\stepcounter{figureNumber}
		\fi
		\caption{AG News}
		\label{fig:mia_agnews}
	\end{subfigure}
	\caption{MIA performance gets worse when the query points become further away from the training points. We define points different from training points but carry similar information as members. In \ref{fig:mia_purchase}, $m$ is the number of missing values. The query is constructed by filling in them with the most likely values. In \ref{fig:mia_celeba}, the point query changes to photos of the same identity who has at least one photo in the training set. In \ref{fig:mia_cifar}, the queries are horizontally flipped images. In \ref{fig:mia_agnews}, $d$ is the Hamming distance to original sentences.}
	\label{fig:mia_attack_performance}
\end{figure*}

\subsection{Range membership inference attack} \label{sec:range_mia_game}
In range membership inference attacks, the goal is to identify if a given \emph{range} contains any training point. We define our range membership inference game, modified from the MI game.
\begin{definition}(\textbf{Range Membership Inference Game}) \label{def:range_mia}
	Let $\pi$ be the data distribution, and let $\mathcal{T}$ be the training algorithm.
	\begin{enumerate}
		\item The \emph{challenger} samples a training dataset $D \longleftarrow \pi$, and trains a model $\theta \longleftarrow \mathcal{T}(D)$.
		\item The \emph{challenger} samples a data record $z_0 \longleftarrow \pi$ from the data distribution, and a training data record $z_1\longleftarrow D$.
		\item The \emph{challenger} flips a fair coin to get the bit $b \in \{0, 1\}$. If $b=1$, the \emph{challenger} samples a range $\mathcal{R}_1$ containing~$z_1$. Otherwise,  \emph{challenger} samples a range $\mathcal{R}_0$ containing $z_0$ and no training points.
		\item The \emph{challenger} sends the target model $\theta$ and the range~$\mathcal{R}_b$ to the \emph{adversary}.
		\item The \emph{adversary} gets access to the data distribution $\pi$ and access to the target model, and outputs a bit $\hat{b} \longleftarrow \mathcal{A}(\theta, \mathcal{R}_b)$.
		\item If $\hat{b}=b$, output 1 (success). Otherwise, output 0.
	\end{enumerate}
\end{definition}

The key difference is that the adversary now receives a \emph{range query} (Step 4) rather than a single data point. We assume that the adversary can sample a set of points from any given range—a reasonable assumption given their ability to sample from the data distribution $\pi$ as in traditional MIAs~\cite{shokri2017membership, ye2022enhanced, zarifzadeh2023low}.

\paragraph{What is a range} A range can be defined by a center, which is a point, a radius representing the size of the range, and a distance function which the radius is defined with. We refer to the center as the query center, the radius as the range size, and the distance function as the range function in this paper. Formally, we can define a range by $\mathcal{R}=\{x':d(x', x) \leq \epsilon\}$, where $x$ is the range center, $d$ is the range function and $\epsilon$ being the range size. One way to visualize a range is to imagine an  $l_2$ ball around a point $x$, replacing the radius and $l_2$ distance with any arbitrary choice of range sizes and functions. Our framework is flexible to accommodate any distance function that preserves a significant amount of the sensitive information. The range function can be spatial (e.g. $l_p$ norms), transformation-based (e.g. geometric transformations), or semantic (e.g., based on user identity). In the experiment section, we will present results with all of these types of range functions. 
Notably, RaMIA reduces to user-level inference~\cite{mahloujifar2021membership, kandpal2023user, liu2021understanding, chen2023face, chen2023slmia} when the range function is defined on a per-user basis.

\paragraph{How to construct a range} In Step 3 of the range membership inference game, the specific procedure for constructing ranges is intentionally left unspecified. This is for flexibility: ranges can be constructed around either in-distribution (ID) or out-of-distribution (OOD) data points for both IN and OUT cases. The details of our range construction methods for experiments are provided in Section~\ref{sec:setup}.

\subsection{Evaluation of RaMIA}
RaMIA is evaluated similarly to MIAs using AUC metrics, but with definitions adapted to the range setting. A range is considered IN if it contains at least one training point and OUT otherwise. Thus, the TPR is defined as the proportion of IN ranges correctly identified by the adversary, and the FPR is the proportion of OUT ranges incorrectly classified as IN. To avoid confusion, we call them (Range) TPR/FPR.

%% file: figure_scripts/purchase_ramia_vs_mia.tex
\begin{tikzpicture}
	\begin{axis}[
		scale=0.9,
		xlabel={FPR}, ylabel={TPR},
		ylabel style={yshift=-0.3cm},
		xmin = 0, xmax = 1,
        ymin = 0, ymax = 1, yscale=1,
		xtick={0.0,0.2,0.4,0.6,0.8,1.0}, xticklabels={0.0,0.2,0.4,0.6,0.8,1.0},
		grid = major, title style={yshift=0.9cm},
		legend style={at={(1.0, 0.0)},anchor=south east, font=\tiny}
		]	

		\addplot[solid, very thick, matplotblue, no marks] table[skip first n=1,x index=0, y index=1, col sep=comma] {"data/purchase/rmia_clean.csv"};
		
		\addplot[solid, very thick, matplotcyan, no marks] table[skip first n=1,x index=0, y index=1, col sep=comma] {"data/purchase/rmia_ad_guess_r10.csv"};
		
		\addplot[solid, very thick, matplotgreen, no marks] table[skip first n=1,x index=0, y index=1, col sep=comma] {"data/purchase/rmia_ad_guess_r100.csv"};
		
		\addplot[solid, very thick, matplotbgreen, no marks] table[skip first n=1,x index=0, y index=1, col sep=comma] {"data/purchase/rmia_ad_guess_r200.csv"};
		
		
		\addplot[densely dotted, thick, black] expression {x};
		
		\legend{MIA (AUC=0.635), MIA (m=10) (AUC=0.632), MIA (m=100) (AUC=0.563), MIA (m=200) (AUC=0.524)}
	\end{axis}
\end{tikzpicture}

%% file: figure_scripts/celeba_ramia_vs_mia.tex
\begin{tikzpicture}
	\begin{axis}[
		scale=0.9,
		xlabel={FPR}, ylabel={TPR},
		ylabel style={yshift=-0.3cm},
		xmin = 0, xmax = 1,
        ymin = 0, ymax = 1, yscale=1,
		xtick={0.0,0.2,0.4,0.6,0.8,1.0}, xticklabels={0.0,0.2,0.4,0.6,0.8,1.0},
		grid = major, title style={yshift=0.9cm},
		legend style={at={(1.0, 0.0)},anchor=south east, font=\tiny}
		]	

		\addplot[solid, very thick, matplotblue, no marks] table[skip first n=1,x index=0, y index=1, col sep=comma] {"data/celeba/rmia_norange_clean.csv"};
		\addplot[solid, very thick, matplotcyan, no marks] table[skip first n=1,x index=0, y index=1, col sep=comma] {"data/celeba/rmia_norange.csv"};
		
		\addplot[densely dotted, thick, black] expression {x};
		
		\legend{MIA (AUC=0.995), MIA (other photos) (AUC=0.560)}
	\end{axis}
\end{tikzpicture}

%% file: figure_scripts/cifar10_matched_vs_unmatched_auc_clean_model.tex
\begin{tikzpicture}
	\begin{axis}[
		scale=0.9,
		xlabel={FPR}, ylabel={TPR},
		xmin = 0, xmax = 1,
        ymin = 0, ymax = 1, yscale=1,
		xtick={0.0,0.2,0.4,0.6,0.8,1.0}, xticklabels={0.0,0.2,0.4,0.6,0.8,1.0},
		grid = major, title style={yshift=0.9cm},
		legend style={at={(1.0, 0.0)},anchor=south east, font=\tiny}
		]	
	
		\addplot[solid, very thick, matplotblue, no marks] table[skip first n=1,x index=0, y index=1, col sep=comma] {"data/cifar/noaug_rmia_norange_base_wrn.csv"};
		\addplot[solid, very thick, matplotcyan, no marks] table[skip first n=1,x index=0, y index=1, col sep=comma] {"data/cifar/noaug_rmia_norange_unmatched_wrn.csv"};

		\addplot[densely dotted, thick, black] expression {x};
		
		\legend{MIA (AUC=0.762), MIA (flipped) (AUC=0.560)}
	\end{axis}
\end{tikzpicture}

%% file: figure_scripts/agnews_ramia_vs_mia.tex
\begin{tikzpicture}
	\begin{axis}[
		scale=0.9,
		xlabel={FPR}, ylabel={TPR},
		ylabel style={yshift=-0.3cm},
		xmin = 0, xmax = 1,
        ymin = 0, ymax = 1, yscale=1,
		xtick={0.0,0.2,0.4,0.6,0.8,1.0}, xticklabels={0.0,0.2,0.4,0.6,0.8,1.0},
		grid = major, title style={yshift=0.9cm},
		legend style={at={(1.0, 0.0)},anchor=south east, font=\tiny}
		]	
		
		\addplot[solid, very thick, matplotblue, no marks] table[skip first n=1,x index=0, y index=1, col sep=comma] {"data/agnews/rmia_k1a1_no_range_clean.csv"};
		\addplot[solid, very thick, matplotcyan, no marks] table[skip first n=1,x index=0, y index=1, col sep=comma] {"data/agnews/norange_rmia_subset.csv"};
		\addplot[solid, very thick, matplotgreen, no marks] table[skip first n=1,x index=0, y index=1, col sep=comma] {"data/agnews/norange_rmia_subset_a5.csv"};
		
		\addplot[densely dotted, thick, black] expression {x};
		
		\legend{MIA (AUC=0.745), MIA (d=2)(AUC=0.609), MIA (d=5) (AUC=0.578)}
	\end{axis}
\end{tikzpicture}

%% file: sections/sec_range_mia.tex
\section{Range membership inference attacks} \label{sec:ramia}
\subsection{(Simple) Hypothesis testing} \label{sec:lrt}
The standard way to tackle the inference game (Def \ref{def:mia}) is to apply statistical hypothesis tests \cite{ye2022enhanced, carlini2022membership}:
\begin{align*}
	H_0&: \text{The given }z \text{ is not a training point }(b=0).\\
	H_1&: \text{The given }z \text{ is a training point }(b=1).
\end{align*}

The likelihood ratio test (LRT) is then conducted
\begin{equation}
	\frac{\probP{\left(\theta|H_1\right)}}{\probP{\left(\theta|H_0\right)}}
\end{equation}

This is usually called "simple" hypothesis testing because each $H$ contains a single hypothesis. The scoring function in the membership inference attack can be considered as an approximation of the likelihood function $\probP{(\cdot)}$. 

\subsection{Composite hypothesis testing}
When extending to range membership inference, the adversary is presented with a set of points rather than a single point. For a given range \(\mathcal{R}_b\), we define the hypotheses as:
\begin{align*}
	H_0&: \text{None of the points in the given range are} \\
	&\text{from the training set. } \forall z \in \mathcal{R}_b: z \notin D.\\
	H_1&: \text{There is at least one point in the given range} \\
	& \text{from the training set. } \exists z \in \mathcal{R}_b \text{ s.t. } z \in D.
\end{align*}
Since it can be intractable to iterate over all points in a given range $\mathcal{R}_b$, we use a set of sampled points $S$ in the given range as its proxy to replace $\mathcal{R}_b$ in the hypotheses above. 

The likelihood ratio in this case is still $	\frac{\probP{\left(\theta|H_1\right)}}{\probP{\left(\theta|H_0\right)}}$. However, the alternative hypothesis $H_1$ is \emph{composite}, representing a union of multiple hypotheses $\bigcup_{z_i \leftarrow S} (z_i \in D)$. Therefore, we need to use statistical methods tailored for composite hypothesis testing. Two commonly used methods are Bayes Factor \cite{jeffreys1939theory} and Generalized Likelihood Ratio Tests (GLRTs) \cite{van1968detection}. 
\paragraph{GLRT} GLRT assumes that the true hypothesis $h^*$ is explicitly present in the composite hypothesis, which, in our case, means the training point is in $S$. This allows us to reduce the composite hypothesis by taking the maximum over $S$: 
\begin{equation} \label{eqn:glrt}
	\probP{\left(\theta|H_1\right)} \Rightarrow	\max_{x \in S} \probP{(\theta|x \in D)}.
\end{equation}

\paragraph{Bayes Factor}
Bayes Factor treats the hypothesis as a random variable with a prior distribution. Each point z in the range is sampled with probability determined by \(\probP{(z)}\), and the composite likelihood is approximated via the expectation:
\begin{equation}
	\probP{\left(\theta|H_1\right)} \Rightarrow \int_{x \in S} \probP{(\theta|x \in D)}\probP{(x)} dx.
\end{equation}

\paragraph{Why both methods fall short in RaMIA}
At first look, GLRT might be an intuitive choice, as it is equivalent to a two-step strategy: search and test. Searching for the points with the highest membership score is conceptually equivalent to identifying the points that are most likely to be training points, and their membership should be indicative of the ranges' membership. However, this assumes the true training point is in every \textbf{sampled} set $S$, which is extremely unlikely. Secondly, this also assumes that we can reliably find the max values in a given range. Since most ranges are large data subspaces, it is very challenging to find the extreme points. Even if the search space can be navigated, search algorithms are likely to return local maxima. 

Similarly, the Bayes Factor approach requires knowledge of the true prior \(\probP{(x)}\), which is generally unknown. 
Both methods also hinge on the absolute correctness of the likelihood values: ideally, the highest likelihood among non-members should be lower than the lowest likelihood among members. However, membership inference attacks are known to be imperfect and particularly unreliable on OOD data~\cite{zarifzadeh2023low}, which may assign high scores erroneously to non-members. This leads to increased Range FPR and reduced AUC, especially when working with sampled data.

\subsection{Our approach: Trimmed averages}
To overcome these limitations, we propose a robust attack strategy based on a modified Bayes Factor approach that employs trimmed averages. Our method begins by assuming every sampled point in $S$ is equally probable. To mitigate the unreliability of MIA scores, we introduce a trimming process that adapts to the nature of the data in the sampling space:

\begin{itemize}
\item \textbf{In-Distribution (ID) Data:} When the sampled points are naturally in-distribution, we trim the lower quantiles (i.e., those with the smallest likelihood values) and average the top samples. This reduces the influence of non-members and random noise.
\item \textbf{Synthetic Data:} When only synthetic data are available, the highest scoring samples are often OOD and prone to false positives. In this case, we trim the top quantiles and average the remaining samples.
\end{itemize}

This strategy is formalized as:
\begin{equation}\label{eqn:strategy}
	\begin{multlined}
		\probP{\left(\theta|H_1\right)} = \text{TrimmedAvg}(S, q_s, q_e; 	\probP)\\
		=\text{Avg}_{x\not \in [q_s, q_e]\text{-th quantiles}} \probP{(\theta|x \in D)},
	\end{multlined}
\end{equation}
where $S$ is the sampled set,  $q_s$ and $q_e$ denote the quantile thresholds between which to be trimmed. For synthetic data, the chance of the top samples being false positives is high, so we set $q_e=100$ to remove the largest points. $q_s$ is a hyperparameter that decreases (trim more) as the quality of sampled points gets worse. For real data, we set $q_s=0$ to remove the smallest points in our aggregation. $q_e$ decreases (trim less) as the number of real samples decreases to offset the high variance due to limited samples available. If we do not trim anything, the formulation reduces to Bayes Factor which assigns equal probability to all points. We provide an empirical comparison in Fig \ref{fig:vs_glrt_bayes} to show that the trimmed means are better suited for RaMIA.

Note that the optimal hyperparameters may vary with different membership signals (e.g., loss values, LiRA scores) since these signals capture different vulnerabilities. For fixed model architectures, range functions, data distributions, and sampling methods, these hyperparameters can be determined using reference models, similar to the offline version of RMIA~\cite{zarifzadeh2023low}. By randomly choosing a reference model as a temporary target model and sweeping through hyperparameters via grid search, one can identify the best values (details in Appendix~\ref{app:attacks}).

\subsection{Range membership inference attack as a framework}
Our proposed RaMIA is not a single attack algorithm but a new inference attack framework. It comprises two key components:

\begin{itemize}
\item \textbf{Sampler:} $\text{Sampler}(\mathcal{R}): \mathcal{R} \rightarrow S$ returns samples within the given range
\item \textbf{Membership Tester:} \text{MIA}(x) is any point-query membership inference algorithm that outputs a membership score, approximating $\probP{(\theta \mid x)}$. This function can be replaced by any existing MIA algorithm.
\end{itemize}
The core idea is to compute a range membership score $\text{RaMIA}(\mathcal{R})$ by robustly aggregating the membership scores of the sampled points $S$ using our trimmed average approach (Eqn.~\ref{eqn:strategy}). This framework leverages existing membership scoring functions while addressing the new challenge of capturing privacy leakage in a broader neighborhood.

\begin{algorithm}[!htb]
	\caption{Computing range membership scores}
	\label{algo:ramia}
	\begin{algorithmic}[1]
		\Require Input range $\mathcal{R}$, sampler $\text{Sample}(\cdot)$, target model $\theta$, membership scoring function $\text{MIA}(\cdot)$.
		\State Sample an attack set:  $S \overset{n}{\longleftarrow} \text{Sample}(\mathcal{R})$;
		\If{samples are real and ID}
		\State Set $q_s=0$, and set $q_e$ by sweeping on reference models;
		\Else
		\State Set $q_e=100$, and set $q_s$ by sweeping on reference models.
		\EndIf
		\State $\text{RaMIA}(\mathcal{R};\theta)=\text{TrimmedAvg}(S, q_s, q_e; \text{MIA})$
	\end{algorithmic}
\end{algorithm}

Importantly, our formulation of RaMIA as a composite hypothesis test—and the subsequent development of the trimmed averages approach—represents a first, yet promising, attempt to tackle this new technical challenge. While there is room for further exploration and refinement, our approach demonstrates that extending the membership inference framework to encompass ranges can capture significant privacy leakage in the vicinity of training data. This represents a core contribution of our work and lays the groundwork for future research in comprehensive privacy auditing.

%% file: sections/sec_experiments.tex
\section{Experiments}
Since the purpose of this paper is to introduce a new concept and framework, the goal of the experiments section is to provide a \textbf{proof-of-concept}.

We experiment on the commonly used Purchase-100 \cite{shokri2017membership}, CelebA \cite{liu2018large}, CIFAR-10 \cite{krizhevsky2009learning} and AG News \cite{zhang2015character} datasets. Details on dataset splits, model training, range construction, and sample acquisition are described in Section \ref{sec:setup}. Since the range membership notion is new, we do not have a prior method to compare with. However, as our aim in introducing this new privacy notion is to enable better and more comprehensive privacy auditing, we compare our RaMIA framework with the de facto privacy auditing framework, MIA. \textbf{Note} that It is crucial to emphasize that our experimental comparison between RaMIA and MIA is not based on the standard MIA notion of membership. In standard MIAs, only the exact training data are considered members. In our experiments, the queries for MIA are the range centers—which are \emph{not} training data by definition. Consequently, if one were to evaluate MIA using the correct MIA privacy notion, the AUC would be close to random guessing (i.e., 0.5). Instead, we use the MIAScore of the range center to solve the range membership inference game, even though this deviates from the standard definition. This allows us to compare the power of both frameworks in identifying queries that leak privacy.

Table \ref{tab:queries} outlines the range queries and point queries used for RaMIA and MIA respectively, while Table \ref{tab:setups} defines the notion of members under each attack setting. In both tables, $x$ represents the original data in the datasets, while $x'$ represents either data with missing values or modified data derived from~$x$. The reason we do not center ranges at the original data $x$ is that, without sufficient prior knowledge, the probability of the attack data exactly matching a training point is extremely low. In practice, similar but not identical data are more likely to be queried. It also acts as a hard case for RaMIA.

\begin{table*}
	\caption{Range queries and point queries used in our experiments for RaMIA and MIA respectively.}
	\label{tab:queries}
	\centering
	\begin{tabular}{@{}ccc@{}}
		\toprule
		Dataset      & Range query                                               & Point query         \\ \midrule
		Purchase-100 & possible data records given the incomplete data $x'$      & mode imputed $x'$ \\
		CelebA       & photos featuring the same person as photo $x'$            & photo $x'$          \\
		CIFAR-10     & transformed versions of image $x'$                        & image $x'$          \\
		AG News      & sentences that are of Hamming distance 8 to sentence $x'$ & sentence $x'$       \\ 
		\bottomrule
	\end{tabular}
\end{table*}

\begin{table*}
	\caption{Definitions of range and point members corresponding to the attack queries in Table \ref{tab:queries}.}
	\label{tab:setups}
	\centering
	\begin{tabular}{ccc}
		\toprule
		Dataset      & Range member if there is at least                            & (Point) member if          \\ \midrule
		Purchase-100 & one training point matches with $x'$ on all unmasked columns & $x'_\text{impu}$ is member \\
		CelebA       & one training image featuring the same person as $x'$         & $x'$ is member     \\
		CIFAR-10     & one version of image $x'$ in the training set                & $x'$ is member      \\
		AG News      & one training sentence within Hamming distance 8 to $x'$      & $x'$ is member  \\ 
		\bottomrule
	\end{tabular}
\end{table*}

\subsection{Setup details} \label{sec:setup}
As mentioned earlier, the range function must derive from the semantics of privacy. Hence, in experiments, we use specific range functions tailored to different data types. For tabular data, we consider missing columns, which is an extreme case of using Euclidean distance on missing columns as a range function ($R=\{x':d(x'_i, x_i) = 0 \land d(x'_j, x_j) < C_j\}$, where $j$'s are masked columns and $i$'s are observable columns, $d$ is Euclidean distance, $C$ is an upper bound for each missing column to make sure $x'_j$ is bounded by the infinity or extreme values (e.g., Age.). For human photos, we use a semantic range function based on the main person featured. For other image data, we use geometric transformations as range functions. For text data, we use (word level) Hamming distance, which is edit or Levenshtein distance that only considers word substitution. The reasons for choosing these range functions have been motivated and explained in earlier parts of this paper.
\subsubsection{Tabular data: Purchase-100}
\begin{itemize}
	\item \textbf{Dataset} Purchase-100 \cite{shokri2017membership} is derived from Kaggle's Acquire Valued Shoppers Challenge \footnote{https://www.kaggle.com/c/acquire-valued-shoppers-challenge/data}. It contains 600 binary features, representing the purchase history of each person. The data is divided into 100 classes. The task is to predict a person's category given the purchase history.
	
	\item \textbf{Models} We train a four-layer multi-layer perceptron (MLP) in PyTorch \cite{paszke2019pytorch} on half of the entire dataset. The hidden layers are of sizes $[1024, 512, 256]$.  All models achieve a test accuracy of $86\%$.
	
	\item \textbf{Construction of ranges}  We simulate the scenario where the attacker has incomplete data (data with missing values). For all training and test data records, we randomly mask k columns. Each row with masked columns is a range query that contains $2^{k}$ possible points. We re-labeled any range that contains at least one training point as "IN".
	
	\item \textbf{Sampling within ranges} Since this dataset contains 600 independent binary features, we do Bernoulli sampling independently for all missing columns. The parameter of the sampler is computed by taking the average value of each column. Due to nature of this dataset, our sampled data can be regarded as ID. We take 19 samples for each range, together with the data obtained by doing mode imputation (fill in the missing values with the modes).
\end{itemize}

\subsubsection{Image data I: CelebA}
\begin{itemize}
	\item \textbf{Dataset} CelebA \cite{liu2018large}, also known as the CelebFaces Attributes dataset, contains  202,599 face images from 10,177 celebrities, each annotated with 40 binary facial features. We construct the members set by only including photos of the first 5090 celebrities. The rest are used to construct the non-members set. For each celebrity in the members set, half of the photos are put into the training set, while the other half goes into the holdout set. 
	
	\item \textbf{Models} We train four-layer convolutional neural networks (CNNs) in PyTorch \cite{paszke2019pytorch} on the training set to predict the facial attributes of any given photo. Our target model has a test accuracy of $87\%$.
	
	\item \textbf{Construction of ranges}  The range function here is semantic, defined by identity. For example, a range query can be "all photos of Alice". Since the identities in the training and non-members set are disjoint, it is easy to construct IN and OUT ranges.
	
	\item \textbf{Sampling within ranges} For each range query, we collect all holdout images sharing the same identity as the range center.
\end{itemize}

\subsubsection{Image data II: CIFAR-10}
\begin{itemize}
	\item \textbf{Dataset} CIFAR-10 \cite{krizhevsky2009learning} is a popular image classification dataset, containing 50,000 training images of size $(32,32,3)$.
	
	\item \textbf{Models} We train WideResNets-28-2 \cite{zagoruyko2016wide} with JAX \cite{jax2018github} on half of the training set of CIFAR-10 using the code from \cite{carlini2022membership}, with and without image augmentations. Our target model achieves a test accuracy of $83\%$ when trained without augmentation, and $92\%$ with augmentation. The train-time augmentation is the composition of random flipping, cropping and random hue. 
	
	\item \textbf{Construction of ranges}  The range function is defined by a set of geometric transformations (e.g., flipping, rotation, cropping). A range query consists of various transformed versions of image $x'$.
	\item \textbf{Sampling within ranges} For each range query, we independently apply 10 image augmentations to the range center. The augmentations include flipping, random rotation, random resizing and cropping, random contrast, brightness, hue, and the composition of them.
\end{itemize}

\begin{figure*}[h]
	\centering
	\begin{subfigure}[b]{0.49\textwidth}
		\centering
		\if\compileFigures1
		\input{figure_scripts/purchase_ramia_vs_mia_clean.tex}
		\else
		\includegraphics[]{fig/\filename-figure\thefigureNumber.pdf}
		\stepcounter{figureNumber}
		\fi
		\caption{Purchase-100 (m=10)}
		\label{fig:attack_purchase}
	\end{subfigure}
	\hfill
	\begin{subfigure}[b]{0.49\textwidth}
		\centering
		\if\compileFigures1
		\input{figure_scripts/celeba_ramia_vs_mia_clean.tex}
		\else
		\includegraphics[]{fig/\filename-figure\thefigureNumber.pdf}
		\stepcounter{figureNumber}
		\fi
		\caption{CelebA}
		\label{fig:attack_celeba}
	\end{subfigure}
	\vskip\baselineskip
	\begin{subfigure}[b]{0.49\textwidth}
		\centering
		\if\compileFigures1
		\input{figure_scripts/cifar10_ramia_vs_mia_clean.tex}
		\else
		\includegraphics[]{fig/\filename-figure\thefigureNumber.pdf}
		\stepcounter{figureNumber}
		\fi
		\caption{CIFAR-10}
		\label{fig:attack_cifar}
	\end{subfigure}
	\hfill
	\begin{subfigure}[b]{0.49\textwidth}
		\centering
		\if\compileFigures1
		\input{figure_scripts/agnews_ramia_vs_mia_clean.tex}
		\else
		\includegraphics[]{fig/\filename-figure\thefigureNumber.pdf}
		\stepcounter{figureNumber}
		\fi
		\caption{AG News (d=5)}
		\label{fig:attack_agnews}
	\end{subfigure}
	\caption{RaMIAs perform better than MIAs on similar but not identical data points to training points that carry overlapping private information. This is the hardest setting for RaMIA where there is only one training point in each IN range, and they are not situtated at the range centers. We will show how RaMIAs will perform better in less restricted scenarios in a later figure. Note that the TPRs/FPRs are the range versions, and the MIA is the approach that uses the range center's MIA score as the range's membership signal, both introduced in Sec \ref{sec:metrics}. }
	\label{fig:attack_performance}
\end{figure*}

\subsubsection{Textual data: AG News}
\begin{itemize}
	\item \textbf{Dataset} We use the popular AG News dataset \cite{zhang2015character}, which is a news collection with four categories of news. We treat it as a text generation dataset, disregarding their labels. It contains 120,000 sentences in the training set.
	
	\item \textbf{Models} We took pretrained GPT-2 \cite{radford2019language} models from Hugging Face's transformers library, and finetuned them on half of AG News' training set with LoRA \cite{hu2021lora} (implemented in Hugging Face's PEFT \cite{peft} library). The finetuning is done for 4 epochs. Our target model achieves a test perplexity of 1.39.
	
	\item \textbf{Construction of ranges}  The range function here is word-level Hamming distance, which can be thought of as the edit distance measured on word level that only allows word substitution. An example of a range query is "all sentences within Hamming distance $d$ to sentence $x$". To construct IN  and OUT ranges, we construct range centers by randomly masking $\alpha$ words from the training and test sentences, before filling in the mask with a pretrained BERT \cite{devlin2018bert} model, so they have a distance of $\alpha$ to the original training/test sentences.
	
	\item \textbf{Sampling within ranges} We mask the range center by $k$ words where $k$ is the Hamming distance specified by the range. Then we use BERT \cite{devlin2018bert} to complete the masks.
\end{itemize}

\subsection{Hyperparameters}
Overall, on Purchase-100, we take 20 samples in every range, and set $q_e=100, q_s=45$. On CIFAR-10, we apply up to 10 distinct transforms, and set $q_e=100, q_s=40$. On AG News, we construct 50 sentences within each range, and set $q_e=100, q_s=20$. On CelebA, each celebrity has a different number of images in the sampling space, ranging from 1 to 18. Since it is hard to standardize the sample size for all ranges, we take all of them. We then set $q_s=0$ and $q_e=25$, which means we are not trimming anything for ranges with very few samples available.

\subsection{Implementation details}
We train 16 models on Purchase-100, and 4 models on CelebA, CIFAR-10, and AG News. Each model is trained on half of its respective dataset, following setups from \cite{carlini2022membership, zarifzadeh2023low}. For all PyTorch models, we use Adam with a learning rate of 0.001. For WideResNets, we use the training code from \cite{carlini2022membership}. AG News models are trained for 4 epochs, and other datasets for 100 epochs. Training is conducted on two Nvidia RTX 3090 GPUs, with AG News taking about 1 hour per epoch and other models less than one hour each.

\subsection{Metrics} \label{sec:metrics}
To evaluate the performance of RaMIA and MIA using the same membership inference backbone, we measure AUCs. The inputs to RaMIA are range queries (as specified in Table \ref{tab:queries}), while the inputs to MIA are the point queries corresponding to range centers. Importantly, the ground truth membership for both RaMIA and MIA is defined according to range membership (see Table \ref{tab:setups}). This means that even though the MIA attack is based on point queries, the true membership is based on whether the range contains at least one training point. If evaluated against the correct MIA privacy notion (which only considers exact training points as members), the AUC would be close to random guessing ($\approx 0.5$). Our experimental setup uses the MIAScore of the range center as a proxy to solve the range membership inference game, allowing us to compare the power of both frameworks in identifying privacy leaking queries. It also illustrates the utility of sampling within the range to construct a more meaningful score under RaMIA's privacy notion. Following \cite{carlini2022membership}, we also report (Range) TPR at small (Range) FPR for both methods (see Table \ref{tab:tpr_fpr}).

\subsection{RaMIAs quantify privacy risks more comprehensively than MIAs} \label{sec:attack_performance}
As we have explained before, data points that are close enough to the training data are out of the scope of membership inference attacks. We observe from Figure \ref{fig:attack_performance} and \ref{fig:attack_performance_lira} that range membership inference attacks are better at identifying those nearby points, and thus providing more comprehensive privacy auditing on all the four datasets we tested. We want to highlight that this is the \textbf{hardest} setting for RaMIA, where each IN range only contains one training point that is also not the range center. We also want to emphasize that the gain is remarkable if we consider how little samples were taken compared to the range sizes. On Purchase-100, there are a total of 1024 candidates, and we take less than 20\% of them. On AG News, there are millions of sentences within a distance of 8. 50 sentences are too little to meaningfully cover anything in the space. Yet, limited samples can lead to noticeable gains, which further shows the current privacy quantification approach is suboptimal and needs a better framework. Due to randomness in sampling, we report the average gain of RaMIA over MIA with standard deviation in Table \ref{tab:aucs}.TPRs at small FPRs are in Table \ref{tab:tpr_fpr}. The improvement in AUC is summarized in Table \ref{tab:aucs}.

\begin{table}[!hb]
	\caption{ Improvement in AUCs after switching from MIA to RaMIA across multiple iterations of random sampling. We do not randomly sample but use all available attack images in CelebA.}
	\label{tab:aucs}
	\centering
	\begin{tabular}{lcccc}
		\toprule
		
		& {Purchase-100} & {CIFAR-10} & {CelebA} & {AG News} \\ 
		\cmidrule{2-5} 								 
		$\Delta \text{AUC/RMIA}$        &	$2.62 \pm 0.04$ 	 		&	$2.13 \pm 0.06$			&	$5.4$			&	$1.20 \pm 0.2$			\\		  			
		$\Delta \text{AUC/LiRA}$        &	$0.90 \pm 0.03$ 	 		&	$ 1.10 \pm 0.00$			&	$4.1$			&	$ 3.80 \pm 0.20 $			\\		  					    		 
		\bottomrule
	\end{tabular}
\end{table}

\paragraph{Relation to user-level inference}
Note that in the CelebA experiment, we use identify information as the semantic range function, making it similar to user-level inference. This further shows that RaMIA is a better and more comprehensive privacy auditing framework. In terms of algorithms, our attack strategy should dominate the simple averaging approach used in prior work \cite{mahloujifar2021membership, kandpal2023user}, since the trimming ratio is optimized (compared to no trimming in simple averaging). The comparison can be found in Fig \ref{fig:celeba_glrt_bayes}. There are other user-level inference algorithms \cite{chen2023face} that train shadow models, which incurs additional computational costs. But these methods usually consider scenarios where user information is the label, e.g. facial and speech recognition systems that predict user ID as their outputs \cite{chen2023face, chen2023slmia}. This encourages their target models to explicitly cluster data based on user information, making inference easier. On the other hand, our target model, a facial attribute classifier, does not use user information anywhere in the training, making the inference harder.

\begin{table*}
	\caption{True Positive Rate under different attacks on different datasets at small false positive rates of $1\%$ and $0.1\%$. MIAs cannot be conducted on incomplete data, so we fill the missing columns with the modes and run the attack on them. Standard deviation over random sampling iterations is reported, except for CelebA, where we use all available candidates. The TPR and FPR are calculated based on the range membership information, as described in Sec \ref{sec:metrics}. For a fair comparison, we should compare RaMIA and MIA based on the same membership testing backbone, e.g. MIA with RMIA versus RaMIA with RMIA. As we have argued, the test queries are unlikely to be exact matches to training points, hence the MIAs are evalauted on range centers, which should be considered non-members. Hence, the TPR values are not indicative of the attack power. This table is only for interested readers who wants to know the attack performance at small FPRs.}
	\label{tab:tpr_fpr}
	\centering
		\begin{tabular}{lcccccccc}
			\toprule
			& \multicolumn{2}{c}{Purchase-100} & \multicolumn{2}{c}{CIFAR-10} & \multicolumn{2}{c}{CelebA} & \multicolumn{2}{c}{AG News} \\ 
			\cmidrule{2-9} 
			TPR@FPR(\%)									 & 1\%          & 0.1\%         & 1\%          & 0.1\%         & 1\%         & 0.1\%        & 1\%         & 0.1\%         \\ 
			\midrule
			\textbf{MIA}        		   &              &               &              &               &             &              &             &               \\
			LOSS               											    &      0        &    0           &     0.92       &     0.02          &      1.86    &    0.31     &    $0.08$        &   $0 $            \\
			RMIA                											&      2.18       &     0.37          &  0.99   &  0.09       &     1.69     &   0.19     &     $0.67$        &   $0.04$               \\
			LiRA 															&      5.20       &     0.02          &  2.12   &  0.57      &     1.68    &   0.24     &     0.68        &  0.00               \\
			\midrule
			\textbf{RaMIA}     &              &               &              &               &             &              &             &               \\
			LOSS                											&    $0  \pm 0$          &    $0  \pm 0$           &    $0.88 \pm 0.06$        &      $0.09\pm 0.03$     &    1.40         &    0.28          &    $1.10 \pm 0.11$       &     $0 \pm 0$          \\
			RMIA                											&    $2.57 \pm 1.58$          &      $0.57 \pm 0.47$      &    $1.41 \pm 0.00$   &  $0.24 \pm 0.00$       &   1.44          &  0.22      &  $0.54\pm 0.12$        &    $0  \pm 0$               \\ 
			LiRA                											&    $2.47 \pm 1.47$          &      $0.50 \pm 0.41$      &    $1.53 \pm 0.00 $   &  $0.58 \pm 0.00$       &  1.10        &  0.02     &  $0.63 \pm 0.13$        &    $0 \pm 0$               \\ 
			\bottomrule
		\end{tabular}
	\end{table*}
	
	\subsection{Factors affecting RaMIA performance} \label{sec:factors}
	\paragraph{Training data density in the range} Due to the nature of the sampling-based approach, the chance of our attack set including a true training point scales linearly with the density of training points in the range. For a fixed sample size, increasing the range without introducing more training points hurts the attack performance because the chance of the attack set including any training point gets diluted, and vice versa. Figure \ref{fig:range_size_cifar} shows that the performance of RaMIA increases when the range becomes larger in the CIFAR-10 experiment. Since the range function in CIFAR-10 is based on image augmentation methods, increasing the range means the attacker applies more distinct augmentation methods, which effectively increases the chance of the attacker obtaining one of the transformed versions of training images seen by the model during training, thus leading to better attack performance. In Figure \ref{fig:attack_celeba}, we conduct the attack assuming the attacker cannot sample any true training images. As a sanity check, we relax this assumption, and Figure \ref{fig:attack_density_members_in_range_celeba} shows that RaMIA performs monotonically better when the density of training images increases from 0\% to 50\%, when the number of samples is fixed.
	
	\begin{figure}[!htb]
		\begin{subfigure}[b]{0.48\textwidth}
			\centering
			\if\compileFigures1
			\input{figure_scripts/celeba_ramia_increase_training_density.tex}
			\else
			\includegraphics[]{fig/\filename-figure\thefigureNumber.pdf}
			\stepcounter{figureNumber}
			\fi
			\caption{RaMIA on CelebA gets better when the training points available for sampling increases. The percentages are the density of training points in the sampling space.}
			\label{fig:attack_density_members_in_range_celeba}
		\end{subfigure}
		\hfill
		\begin{subfigure}[b]{0.48\textwidth}
			\centering
			\if\compileFigures1
			\input{figure_scripts/cifar10_ramia_k.tex}
			\else
			\includegraphics[]{fig/\filename-figure\thefigureNumber.pdf}
			\stepcounter{figureNumber}
			\fi
			\caption{RaMIA performs better on CIFAR-10 when the range size increases. The size is equal to the number of distinct transformations ($n$) applied to images.\\}
			\label{fig:range_size_cifar}
		\end{subfigure}
		\caption{Attack performance increases in tandem with training point density.}
		\label{fig:attack_performance_range_size}
	\end{figure}
	
	\begin{figure}[!htb]
		\begin{subfigure}[b]{0.48\textwidth}
			\centering
			\if\compileFigures1
			\input{figure_scripts/purchase_l1.tex}
			\else
			\includegraphics[]{fig/\filename-figure\thefigureNumber.pdf}
			\stepcounter{figureNumber}
			\fi
			\caption{Purchase-100 with $\ell_1$ range functions}
			\label{fig:purchase_l1}
		\end{subfigure}
		\hfill
		\begin{subfigure}[b]{0.48\textwidth}
			\centering
			\if\compileFigures1
			\input{figure_scripts/cifar10_l2.tex}
			\else
			\includegraphics[]{fig/\filename-figure\thefigureNumber.pdf}
			\stepcounter{figureNumber}
			\fi
			\caption{CIFAR-10 with $\ell_2$ range functions}
			\label{fig:cifar_l2}
		\end{subfigure}
		\caption{Attack performance decreases when increasing range sizes $d$ without introducing more training members to ranges, which effectively reduces the training data density in the range.}
		\label{fig:attack_performance_range_size_decrease}
	\end{figure}
	
	\begin{figure*}[h!]
		\centering
		\begin{subfigure}[b]{0.32\textwidth}
			\if\compileFigures1
			\input{figure_scripts/purchase_score_corr.tex}
			\else
			\includegraphics[]{fig/\filename-figure\thefigureNumber.pdf}
			\stepcounter{figureNumber}
			\fi
			\caption{Purchase-100}
			\label{fig:purchase_corr}
		\end{subfigure}
		\hfill
		\begin{subfigure}[b]{0.31\textwidth}
			\if\compileFigures1
			\input{figure_scripts/cifar10_score_corr.tex}
			\else
			\includegraphics[]{fig/\filename-figure\thefigureNumber.pdf}
			\stepcounter{figureNumber}
			\fi
			\caption{CIFAR-10}
			\label{fig:cifar_corr}
		\end{subfigure}
		\hfill
		\begin{subfigure}[b]{0.31\textwidth}
			\if\compileFigures1
			\input{figure_scripts/agnews_score_corr.tex}
			\else
			\includegraphics[]{fig/\filename-figure\thefigureNumber.pdf}
			\stepcounter{figureNumber}
			\fi
			\caption{AG News}
			\label{fig:agnews_corr}
		\end{subfigure}
		\caption{ Correlation between the percentile of RaMIA and MIA scores of members among non-members. The larger the percentile is, the more non-members the member dominates, and the more likely for it to be classified as IN. The Pearson correlation coefficients are provided for each plot. It shows that the vulnerability to MIA is positively correlated with that to RaMIA. }
		\label{fig:vulnerablility_corr}
	\end{figure*}
	
	We also observe a monotonic decrease in RaMIA performance as the range size increases without introducing true members (Fig \ref{fig:attack_performance_range_size_decrease}). We use L1 distance as the range function for Purchase-100, and L2 distance for CIFAR-10, and construct ranges around each data point in MIA's evaluation set. We ensure that each IN range only contains one training point, which is the range center, and that each OUT range does not contain any training point. 
	
	\paragraph{Susceptibility to MIAs and RaMIAs is correlated} Ranges containing training points that are susceptible to MIAs are also more susceptible to RaMIAs. Researchers have previously discovered that machine learning models memorize duplicate data more \cite{lee2021deduplicating, carlini2022quantifying}. In our CelebA dataset, each celebrity has a different number of photos in the training set, which can be thought of as each identity having different levels of duplication in the training set. Similar to the insights from MIAs, we also observe that identities that have more training images, i.e. higher duplication rate, are more susceptible to RaMIA. Figure \ref{fig:celeba_duplicates} shows the relationship between the percentile of each range's RaMIA score within non-members' RaMIA scores and the duplication rate. Generally speaking, identities that have more training photos are more prone to RaMIAs. Similarly, correlation can be observed on the other three datasets in our experiments, where the training points' RaMIA score percentiles among non-members are positively correlated with their MIA score percentiles (Fig \ref{fig:vulnerablility_corr}).
	
	\begin{figure}[!htb]
		\centering
		\if\compileFigures1
		\input{figure_scripts/celeba_boxplot.tex}
		\else
		\includegraphics[width=0.4\textwidth]{fig/\filename-figure\thefigureNumber.pdf}
		\stepcounter{figureNumber}
		\fi
		\caption{Number of photos from the same celebrity in the training set affects the identifiability of the range.}
		\label{fig:celeba_duplicates}
	\end{figure}
	
	\subsection{Mismatched training and attack data hurts attack performance} \label{sec:mismatch}
	Figure \ref{fig:mia_cifar} shows that MIA underestimates the privacy risk when the augmentation used in training and attacking differs. This should be alarming as many people audit the privacy risk of image classifiers with original images, when the classifiers are often trained with a composition of augmentations. Many transformations, such as color jittering and affine transformations, always produce different final images. Other commonly used augmentation methods, such as random cropping, introduce more randomness to the pipeline. Hence, it is almost certain that the original images are never seen by the model. Therefore, we should use RaMIA for a better auditing result (Figure \ref{fig:attack_cifar}). 
	
	
	\subsection{RaMIA on redacted data} \label{sec:redacted_data}
	Many large language models (LLMs) are trained with sensitive textual data. Some of the data with sensitive information redacted might be publicly available. Similar to our experiment with data with missing values, we can apply RaMIA to redacted data to identify which of them are used to train a target LLM. Accurately identifying the redacted sentences paves the way for potentially better data extraction and reconstruction attacks. Figure \ref{fig:attack_agnews_pii} shows the results. In this experiment, we use spaCy \cite{Honnibal_spaCy_Industrial-strength_Natural_2020} to mask peoples' names to simulate the masking of personally identifiable information (PII). We then generate 10 possible sentences for each masked sentence using BERT and conduct RaMIA. The MIA performance is the average attack performance over all 10 possible sentences. The performance gap is smaller compared to that in Figure \ref{fig:attack_agnews}. The reason might be that BERT fails to produce diverse PII completions, making all candidate sentences similar to each other, and reducing the power of RaMIA.
	
	\begin{figure}[!htb]
		\centering
		\if\compileFigures1
		\input{figure_scripts/agnews_ramia_vs_mia_pii.tex}
		\else
		\includegraphics[width=0.4\textwidth]{fig/\filename-figure\thefigureNumber.pdf}
		\stepcounter{figureNumber}
		\fi
		\caption{RaMIA on a subset of AG News where names are redacted.}
		\label{fig:attack_agnews_pii}
	\end{figure}

%% file: figure_scripts/purchase_ramia_vs_mia_clean.tex
\begin{tikzpicture}
	\begin{axis}[
		scale=0.9,
		xlabel={(Range) FPR}, ylabel={(Range) TPR},
		ylabel style={yshift=-0.3cm},
		xmin = 0, xmax = 1,
        ymin = 0, ymax = 1, yscale=1,
		xtick={0.0,0.2,0.4,0.6,0.8,1.0}, xticklabels={0.0,0.2,0.4,0.6,0.8,1.0},
		grid = major, title style={yshift=0.9cm},
		legend style={at={(1.0, 0.0)},anchor=south east, font=\tiny}
		]

		\addplot[solid, thick, matplotblue, no marks] table[skip first n=1,x index=0, y index=1, col sep=comma] {"data/purchase/rmia_ad_guess_r10.csv"};
		\addplot[solid, very thick, matplotred, no marks] table[skip first n=1,x index=0, y index=1, col sep=comma] {"data/purchase/range_rmia_bottom6_r10.csv"};
		
		\addplot[densely dotted, thick, black] expression {x};
		
		\legend{MIA (AUC=0.632), RaMIA (AUC=0.657)}
	\end{axis}
\end{tikzpicture}

%% file: figure_scripts/celeba_ramia_vs_mia_clean.tex
\begin{tikzpicture}
	\begin{axis}[
		scale=0.9,
		xlabel={(Range) FPR}, ylabel={(Range) TPR},
		ylabel style={yshift=-0.3cm},
		xmin = 0, xmax = 1,
        ymin = 0, ymax = 1, yscale=1,
		xtick={0.0,0.2,0.4,0.6,0.8,1.0}, xticklabels={0.0,0.2,0.4,0.6,0.8,1.0},
		grid = major, title style={yshift=0.9cm},
		legend style={at={(1.0, 0.0)},anchor=south east, font=\tiny}
		]	



		\addplot[solid, very thick, matplotblue, no marks] table[skip first n=1,x index=0, y index=1, col sep=comma] {"data/celeba/rmia_norange.csv"};
	\addplot[solid, very thick, matplotred, no marks] table[skip first n=1,x index=0, y index=1, col sep=comma] {"data/celeba/range_rmia_k15.csv"};
		
		\addplot[densely dotted, thick, black] expression {x};
		
		\legend{MIA (AUC=0.560), RaMIA (AUC=0.614)}
	\end{axis}
\end{tikzpicture}

%% file: figure_scripts/cifar10_ramia_vs_mia_clean.tex
\begin{tikzpicture}
	\begin{axis}[
		scale=0.9,
		xlabel={(Range) FPR}, ylabel={(Range) TPR},
		ylabel style={yshift=-0.3cm},
		xmin = 0, xmax = 1,
        ymin = 0, ymax = 1, yscale=1,
		xtick={0.0,0.2,0.4,0.6,0.8,1.0}, xticklabels={0.0,0.2,0.4,0.6,0.8,1.0},
		grid = major, title style={yshift=0.9cm},
		legend style={at={(1.0, 0.0)},anchor=south east, font=\tiny}
		]	

		\addplot[solid, very thick, matplotblue, no marks] table[skip first n=1,x index=0, y index=1, col sep=comma] {"data/cifar/aug_rmia_norange_wrn.csv"};
		\addplot[solid, very thick, matplotred, no marks] table[skip first n=1,x index=0, y index=1, col sep=comma] {"data/cifar/aug_rmia_range_wrn.csv"};
		
		\addplot[densely dotted, thick, black] expression {x};
		
		\legend{MIA (AUC=0.601), RaMIA (AUC=0.621)}
	\end{axis}
\end{tikzpicture}

%% file: figure_scripts/agnews_ramia_vs_mia_clean.tex
\begin{tikzpicture}
	\begin{axis}[
		scale=0.9,
		xlabel={(Range) FPR}, ylabel={(Range) TPR},
		ylabel style={yshift=-0.3cm},
		xmin = 0, xmax = 1,
        ymin = 0, ymax = 1, yscale=1,
		xtick={0.0,0.2,0.4,0.6,0.8,1.0}, xticklabels={0.0,0.2,0.4,0.6,0.8,1.0},
		grid = major, title style={yshift=0.9cm},
		legend style={at={(1.0, 0.0)},anchor=south east, font=\tiny}
		]	

		\addplot[solid, very thick, matplotblue, no marks] table[skip first n=1,x index=0, y index=1, col sep=comma] {"data/agnews/norange_rmia_subset_a5.csv"};
	
		\addplot[solid, very thick, matplotred, no marks] table[skip first n=1,x index=0, y index=1, col sep=comma] {"data/agnews/range_rmia_subset_a5.csv"};
		
		\addplot[densely dotted, thick, black] expression {x};
		
		\legend{MIA (AUC=0.578), RaMIA (AUC=0.599)}
	\end{axis}
\end{tikzpicture}

%% file: figure_scripts/celeba_ramia_increase_training_density.tex
\begin{tikzpicture}
	\begin{axis}[
		scale=0.7,
		xlabel={(Range) FPR}, ylabel={(Range) TPR},
		ylabel style={yshift=-0.3cm},
		xmin = 0, xmax = 1,
        ymin = 0, ymax = 1, yscale=1,
		xtick={0.0,0.2,0.4,0.6,0.8,1.0}, xticklabels={0.0,0.2,0.4,0.6,0.8,1.0},
		grid = major, title style={yshift=0.9cm},
		legend style={at={(1.0, 0.0)},anchor=south east, font=\tiny}
		]	

		\addplot[solid,  very thick, matplotblue, no marks] table[skip first n=1,x index=0, y index=1, col sep=comma] {"data/celeba/range_rmia_k15.csv"};
		\addplot[solid,  very thick, matplotcyan, no marks] table[skip first n=1,x index=0, y index=1, col sep=comma] {"data/celeba/range_rmia_mix0.166667.csv"};
		\addplot[solid,  very thick, matplotgreen, no marks] table[skip first n=1,x index=0, y index=1, col sep=comma] {"data/celeba/range_rmia_mix0.285714.csv"};
		\addplot[solid,  very thick, matplotbgreen, no marks] table[skip first n=1,x index=0, y index=1, col sep=comma] {"data/celeba/range_rmia_mix0.375000.csv"};
		\addplot[solid, very thick, matplotpink, no marks] table[skip first n=1,x index=0, y index=1, col sep=comma] {"data/celeba/range_rmia_mix0.444444.csv"};		
		\addplot[solid, very thick, matplotred, no marks] table[skip first n=1,x index=0, y index=1, col sep=comma] {"data/celeba/range_rmia_mix0.500000.csv"};
		
		\addplot[densely dotted, thick, black] expression {x};
		
		\legend{0\% (AUC=0.614), 16.7\% (AUC=0.712), 28.6\% (AUC=0.797),  37.5\% (AUC=0.854), 44.4\% (AUC=0.887), 50.0\% (AUC=0.936)}
	\end{axis}
\end{tikzpicture}

%% file: figure_scripts/cifar10_ramia_k.tex
\begin{tikzpicture}
	\begin{axis}[
		scale=0.7,
		xlabel={(Range) FPR}, ylabel={(Range) TPR},
		ylabel style={yshift=-0.3cm},
		xmin = 0, xmax = 1,
        ymin = 0, ymax = 1, yscale=1,
		xtick={0.0,0.2,0.4,0.6,0.8,1.0}, xticklabels={0.0,0.2,0.4,0.6,0.8,1.0},
		grid = major, title style={yshift=0.9cm},
		legend style={at={(1.0, 0.0)},anchor=south east, font=\tiny}
		]	

		\addplot[solid,  very thick, matplotblue, no marks] table[skip first n=1,x index=0, y index=1, col sep=comma] {"data/cifar/aug_rmia_range_k1.csv"};
		\addplot[solid,  very thick,matplotgreen, no marks] table[skip first n=1,x index=0, y index=1, col sep=comma] {"data/cifar/aug_rmia_range_k4.csv"};
		\addplot[solid,  very thick, matplotbgreen , no marks] table[skip first n=1,x index=0, y index=1, col sep=comma] {"data/cifar/aug_rmia_range_k7.csv"};
		\addplot[solid, very thick, matplotred, no marks] table[skip first n=1,x index=0, y index=1, col sep=comma] {"data/cifar/aug_rmia_range_wrn.csv"};
		
		\addplot[densely dotted, thick, black] expression {x};
		
		\legend{n=1 (AUC=0.500), n=4 (AUC=0.578), n=7 (AUC=0.605), n=10 (AUC=0.621)}

	\end{axis}
\end{tikzpicture}

%% file: figure_scripts/purchase_l1.tex
\begin{tikzpicture}
	\begin{axis}[
		scale=0.7,
		xlabel={(Range) FPR}, ylabel={(Range) TPR},
		ylabel style={yshift=-0.3cm},
		xmin = 0, xmax = 1,
        ymin = 0, ymax = 1, yscale=1,
		xtick={0.0,0.2,0.4,0.6,0.8,1.0}, xticklabels={0.0,0.2,0.4,0.6,0.8,1.0},
		grid = major, title style={yshift=0.9cm},
		legend style={at={(1.0, 0.0)},anchor=south east, font=\tiny}
		]	

		\addplot[solid,  very thick, matplotblue, no marks] table[skip first n=1,x index=0, y index=1, col sep=comma] {"data/purchase/l1/range_rmia_d10.csv"};

		\addplot[solid,  very thick,matplotgreen, no marks] table[skip first n=1,x index=0, y index=1, col sep=comma] {"data/purchase/l1/range_rmia_d20.csv"};

		\addplot[solid, very thick, matplotred, no marks] table[skip first n=1,x index=0, y index=1, col sep=comma] {"data/purchase/l1/range_rmia_d50.csv"};
		
		\addplot[densely dotted, thick, black] expression {x};

		\legend{d=10 (AUC=0.677), d=20 (AUC=0.672), d=50 (AUC=0.621)}
	\end{axis}
\end{tikzpicture}

%% file: figure_scripts/cifar10_l2.tex
\begin{tikzpicture}
	\begin{axis}[
		scale=0.7,
		xlabel={(Range) FPR}, ylabel={(Range) TPR},
		ylabel style={yshift=-0.3cm},
		xmin = 0, xmax = 1,
        ymin = 0, ymax = 1, yscale=1,
		xtick={0.0,0.2,0.4,0.6,0.8,1.0}, xticklabels={0.0,0.2,0.4,0.6,0.8,1.0},
		grid = major, title style={yshift=0.9cm},
		legend style={at={(1.0, 0.0)},anchor=south east, font=\tiny}
		]	

		\addplot[solid,  very thick, matplotblue, no marks] table[skip first n=1,x index=0, y index=1, col sep=comma] {"data/cifar/l2/rmia_d2.csv"};

		\addplot[solid,  very thick,matplotgreen, no marks] table[skip first n=1,x index=0, y index=1, col sep=comma] {"data/cifar/l2/rmia_d5.csv"};
		\addplot[solid, very thick, matplotred, no marks] table[skip first n=1,x index=0, y index=1, col sep=comma] {"data/cifar/l2/rmia_d10.csv"};
		
		\addplot[densely dotted, thick, black] expression {x};
		
		\legend{d=2 (AUC=0.771), d=5 (AUC=0.721), d=10 (AUC=0.654)}
	\end{axis}
\end{tikzpicture}

%% file: figure_scripts/purchase_score_corr.tex
\begin{tikzpicture}

\begin{axis}[
		scale=0.7,
		tick align=outside,
		tick pos=left,
		x grid style={black},
		xmin=-0.02123, xmax=1.04863,
		xtick style={color=black},
		y grid style={black},
		ymin=-0.03635, ymax=1.04935,
		ytick style={color=black},
		xlabel={MIA score percentile},
		ylabel={RaMIA score percentile},
		legend style={at={(1,0)}, anchor=south east},
		]
		\addlegendentry{Corr=0.66}
		\addplot [draw=matplotblue, fill=matplotblue, mark=*, only marks, mark size=0.3] table[skip first n=1,x index=0, y index=1, col sep=comma]{"data/purchase/corr.csv"};
	\end{axis}

\end{tikzpicture}

%% file: figure_scripts/cifar10_score_corr.tex
\begin{tikzpicture}
	
	\begin{axis}[
		scale=0.7,
		tick align=outside,
		tick pos=left,
		x grid style={black},
		xmin=-0.02123, xmax=1.04863,
		xtick style={color=black},
		y grid style={black},
		ymin=-0.03635, ymax=1.04935,
		ytick style={color=black},
		xlabel={MIA score percentile},
		legend style={at={(1,0)}, anchor=south east},
		]
		\addlegendentry{Corr=0.74}
		\addplot [draw=matplotblue, fill=matplotblue, mark=*, only marks, mark size=0.3] table[skip first n=1,x index=0, y index=1, col sep=comma]{"data/cifar/corr.csv"};
	\end{axis}
	
\end{tikzpicture}

%% file: figure_scripts/agnews_score_corr.tex
\begin{tikzpicture}

\begin{axis}[
	scale=0.7,
	tick align=outside,
	tick pos=left,
	x grid style={black},
	xmin=-0.02123, xmax=1.04863,
	xtick style={color=black},
	y grid style={black},
	ymin=-0.03635, ymax=1.04935,
	ytick style={color=black},
	xlabel={MIA score percentile},
	legend style={at={(1,0)}, anchor=south east},
	]
	\addlegendentry{Corr=0.80}
	\addplot [draw=matplotblue, fill=matplotblue, mark=*, only marks, mark size=0.3] table[skip first n=1,x index=0, y index=1, col sep=comma]{"data/agnews/corr.csv"};
\end{axis}

\end{tikzpicture}

%% file: figure_scripts/celeba_boxplot.tex
\begin{tikzpicture}

\begin{axis}[
tick pos=both,
ylabel={Percentile of RaMIA scores among out-ranges},
xmin=0.35, xmax=17.85,
xlabel={Number of training images with the same identities},
ymin=-0.0495676100628931, ymax=1.04956761006289,
ytick={-0.2,0,0.2,0.4,0.6,0.8,1,1.2},
yticklabels={
  \(\displaystyle {\ensuremath{-}0.2}\),
  \(\displaystyle {0.0}\),
  \(\displaystyle {0.2}\),
  \(\displaystyle {0.4}\),
  \(\displaystyle {0.6}\),
  \(\displaystyle {0.8}\),
  \(\displaystyle {1.0}\),
  \(\displaystyle {1.2}\)
}
]
\path [draw=black, fill=matplotblue]
(axis cs:0.75,0.336674528301887)
--(axis cs:1.25,0.336674528301887)
--(axis cs:1.25,0.93656643081761)
--(axis cs:0.75,0.93656643081761)
--(axis cs:0.75,0.336674528301887)
--cycle;
\addplot [black]
table {%
1 0.336674528301887
1 0.0119889937106918
};
\addplot [black]
table {%
1 0.93656643081761
1 0.999606918238994
};
\addplot [black]
table {%
0.875 0.0119889937106918
1.125 0.0119889937106918
};
\addplot [black]
table {%
0.875 0.999606918238994
1.125 0.999606918238994
};
\path [draw=black, fill=matplotblue]
(axis cs:1.75,0.0437303459119497)
--(axis cs:2.25,0.0437303459119497)
--(axis cs:2.25,0.851906446540881)
--(axis cs:1.75,0.851906446540881)
--(axis cs:1.75,0.0437303459119497)
--cycle;
\addplot [black]
table {%
2 0.0437303459119497
2 0.000393081761006289
};
\addplot [black]
table {%
2 0.851906446540881
2 0.999606918238994
};
\addplot [black]
table {%
1.875 0.000393081761006289
2.125 0.000393081761006289
};
\addplot [black]
table {%
1.875 0.999606918238994
2.125 0.999606918238994
};
\path [draw=black, fill=matplotblue]
(axis cs:2.75,0.15123820754717)
--(axis cs:3.25,0.15123820754717)
--(axis cs:3.25,0.842423349056604)
--(axis cs:2.75,0.842423349056604)
--(axis cs:2.75,0.15123820754717)
--cycle;
\addplot [black]
table {%
3 0.15123820754717
3 0.00373427672955975
};
\addplot [black]
table {%
3 0.842423349056604
3 0.999606918238994
};
\addplot [black]
table {%
2.875 0.00373427672955975
3.125 0.00373427672955975
};
\addplot [black]
table {%
2.875 0.999606918238994
3.125 0.999606918238994
};
\path [draw=black, fill=matplotblue]
(axis cs:3.75,0.299626572327044)
--(axis cs:4.25,0.299626572327044)
--(axis cs:4.25,0.864878144654088)
--(axis cs:3.75,0.864878144654088)
--(axis cs:3.75,0.299626572327044)
--cycle;
\addplot [black]
table {%
4 0.299626572327044
4 0.00511006289308176
};
\addplot [black]
table {%
4 0.864878144654088
4 0.999606918238994
};
\addplot [black]
table {%
3.875 0.00511006289308176
4.125 0.00511006289308176
};
\addplot [black]
table {%
3.875 0.999606918238994
4.125 0.999606918238994
};
\path [draw=black, fill=matplotblue]
(axis cs:4.75,0.280365566037736)
--(axis cs:5.25,0.280365566037736)
--(axis cs:5.25,0.839033018867924)
--(axis cs:4.75,0.839033018867924)
--(axis cs:4.75,0.280365566037736)
--cycle;
\addplot [black]
table {%
5 0.280365566037736
5 0.0145440251572327
};
\addplot [black]
table {%
5 0.839033018867924
5 0.995086477987421
};
\addplot [black]
table {%
4.875 0.0145440251572327
5.125 0.0145440251572327
};
\addplot [black]
table {%
4.875 0.995086477987421
5.125 0.995086477987421
};
\path [draw=black, fill=matplotblue]
(axis cs:5.75,0.381682389937107)
--(axis cs:6.25,0.381682389937107)
--(axis cs:6.25,0.90939465408805)
--(axis cs:5.75,0.90939465408805)
--(axis cs:5.75,0.381682389937107)
--cycle;
\addplot [black]
table {%
6 0.381682389937107
6 0.036753144654088
};
\addplot [black]
table {%
6 0.90939465408805
6 0.999606918238994
};
\addplot [black]
table {%
5.875 0.036753144654088
6.125 0.036753144654088
};
\addplot [black]
table {%
5.875 0.999606918238994
6.125 0.999606918238994
};
\path [draw=black, fill=matplotblue]
(axis cs:6.75,0.46747248427673)
--(axis cs:7.25,0.46747248427673)
--(axis cs:7.25,0.876572327044025)
--(axis cs:6.75,0.876572327044025)
--(axis cs:6.75,0.46747248427673)
--cycle;
\addplot [black]
table {%
7 0.46747248427673
7 0.0218160377358491
};
\addplot [black]
table {%
7 0.876572327044025
7 0.997051886792453
};
\addplot [black]
table {%
6.875 0.0218160377358491
7.125 0.0218160377358491
};
\addplot [black]
table {%
6.875 0.997051886792453
7.125 0.997051886792453
};
\path [draw=black, fill=matplotblue]
(axis cs:7.75,0.38782429245283)
--(axis cs:8.25,0.38782429245283)
--(axis cs:8.25,0.844487028301887)
--(axis cs:7.75,0.844487028301887)
--(axis cs:7.75,0.38782429245283)
--cycle;
\addplot [black]
table {%
8 0.38782429245283
8 0.0369496855345912
};
\addplot [black]
table {%
8 0.844487028301887
8 0.995872641509434
};
\addplot [black]
table {%
7.875 0.0369496855345912
8.125 0.0369496855345912
};
\addplot [black]
table {%
7.875 0.995872641509434
8.125 0.995872641509434
};
\path [draw=black, fill=matplotblue]
(axis cs:8.75,0.337116745283019)
--(axis cs:9.25,0.337116745283019)
--(axis cs:9.25,0.813630110062893)
--(axis cs:8.75,0.813630110062893)
--(axis cs:8.75,0.337116745283019)
--cycle;
\addplot [black]
table {%
9 0.337116745283019
9 0.025746855345912
};
\addplot [black]
table {%
9 0.813630110062893
9 0.992334905660377
};
\addplot [black]
table {%
8.875 0.025746855345912
9.125 0.025746855345912
};
\addplot [black]
table {%
8.875 0.992334905660377
9.125 0.992334905660377
};
\path [draw=black, fill=matplotblue]
(axis cs:9.75,0.379520440251572)
--(axis cs:10.25,0.379520440251572)
--(axis cs:10.25,0.839033018867924)
--(axis cs:9.75,0.839033018867924)
--(axis cs:9.75,0.379520440251572)
--cycle;
\addplot [black]
table {%
10 0.379520440251572
10 0.0676100628930818
};
\addplot [black]
table {%
10 0.839033018867924
10 0.999606918238994
};
\addplot [black]
table {%
9.875 0.0676100628930818
10.125 0.0676100628930818
};
\addplot [black]
table {%
9.875 0.999606918238994
10.125 0.999606918238994
};
\path [draw=black, fill=matplotblue]
(axis cs:10.75,0.445705581761006)
--(axis cs:11.25,0.445705581761006)
--(axis cs:11.25,0.832891116352201)
--(axis cs:10.75,0.832891116352201)
--(axis cs:10.75,0.445705581761006)
--cycle;
\addplot [black]
table {%
11 0.445705581761006
11 0.029874213836478
};
\addplot [black]
table {%
11 0.832891116352201
11 0.999606918238994
};
\addplot [black]
table {%
10.875 0.029874213836478
11.125 0.029874213836478
};
\addplot [black]
table {%
10.875 0.999606918238994
11.125 0.999606918238994
};
\path [draw=black, fill=matplotblue]
(axis cs:11.75,0.431505503144654)
--(axis cs:12.25,0.431505503144654)
--(axis cs:12.25,0.846452437106918)
--(axis cs:11.75,0.846452437106918)
--(axis cs:11.75,0.431505503144654)
--cycle;
\addplot [black]
table {%
12 0.431505503144654
12 0.0349842767295597
};
\addplot [black]
table {%
12 0.846452437106918
12 0.999606918238994
};
\addplot [black]
table {%
11.875 0.0349842767295597
12.125 0.0349842767295597
};
\addplot [black]
table {%
11.875 0.999606918238994
12.125 0.999606918238994
};
\path [draw=black, fill=matplotblue]
(axis cs:12.75,0.502014544025157)
--(axis cs:13.25,0.502014544025157)
--(axis cs:13.25,0.820656446540881)
--(axis cs:12.75,0.820656446540881)
--(axis cs:12.75,0.502014544025157)
--cycle;
\addplot [black]
table {%
13 0.502014544025157
13 0.0294811320754717
};
\addplot [black]
table {%
13 0.820656446540881
13 0.994300314465409
};
\addplot [black]
table {%
12.875 0.0294811320754717
13.125 0.0294811320754717
};
\addplot [black]
table {%
12.875 0.994300314465409
13.125 0.994300314465409
};
\path [draw=black, fill=matplotblue]
(axis cs:13.75,0.462755503144654)
--(axis cs:14.25,0.462755503144654)
--(axis cs:14.25,0.814023191823899)
--(axis cs:13.75,0.814023191823899)
--(axis cs:13.75,0.462755503144654)
--cycle;
\addplot [black]
table {%
14 0.462755503144654
14 0.0721305031446541
};
\addplot [black]
table {%
14 0.814023191823899
14 0.995086477987421
};
\addplot [black]
table {%
13.875 0.0721305031446541
14.125 0.0721305031446541
};
\addplot [black]
table {%
13.875 0.995086477987421
14.125 0.995086477987421
};
\path [draw=black, fill=matplotblue]
(axis cs:14.75,0.488502358490566)
--(axis cs:15.25,0.488502358490566)
--(axis cs:15.25,0.832055817610063)
--(axis cs:14.75,0.832055817610063)
--(axis cs:14.75,0.488502358490566)
--cycle;
\addplot [black]
table {%
15 0.488502358490566
15 0.147405660377358
};
\addplot [black]
table {%
15 0.832055817610063
15 0.993514150943396
};
\addplot [black]
table {%
14.875 0.147405660377358
15.125 0.147405660377358
};
\addplot [black]
table {%
14.875 0.993514150943396
15.125 0.993514150943396
};
\path [draw=black, fill=matplotblue]
(axis cs:15.75,0.516951650943396)
--(axis cs:16.25,0.516951650943396)
--(axis cs:16.25,0.829107704402516)
--(axis cs:15.75,0.829107704402516)
--(axis cs:15.75,0.516951650943396)
--cycle;
\addplot [black]
table {%
16 0.516951650943396
16 0.0617138364779874
};
\addplot [black]
table {%
16 0.829107704402516
16 0.999017295597484
};
\addplot [black]
table {%
15.875 0.0617138364779874
16.125 0.0617138364779874
};
\addplot [black]
table {%
15.875 0.999017295597484
16.125 0.999017295597484
};
\path [draw=black, fill=matplotblue]
(axis cs:16.75,0.746462264150943)
--(axis cs:17.25,0.746462264150943)
--(axis cs:17.25,0.862814465408805)
--(axis cs:16.75,0.862814465408805)
--(axis cs:16.75,0.746462264150943)
--cycle;
\addplot [black]
table {%
17 0.746462264150943
17 0.746462264150943
};
\addplot [black]
table {%
17 0.862814465408805
17 0.916077044025157
};
\addplot [black]
table {%
16.875 0.746462264150943
17.125 0.746462264150943
};
\addplot [black]
table {%
16.875 0.916077044025157
17.125 0.916077044025157
};
\addplot [matplotcyan]
table {%
0.75 0.699685534591195
1.25 0.699685534591195
};
\addplot [matplotcyan]
table {%
1.75 0.307979559748428
2.25 0.307979559748428
};
\addplot [matplotcyan]
table {%
2.75 0.523093553459119
3.25 0.523093553459119
};
\addplot [matplotcyan]
table {%
3.75 0.626572327044025
4.25 0.626572327044025
};
\addplot [matplotcyan]
table {%
4.75 0.588443396226415
5.25 0.588443396226415
};
\addplot [matplotcyan]
table {%
5.75 0.748231132075472
6.25 0.748231132075472
};
\addplot [matplotcyan]
table {%
6.75 0.66312893081761
7.25 0.66312893081761
};
\addplot [matplotcyan]
table {%
7.75 0.580188679245283
8.25 0.580188679245283
};
\addplot [matplotcyan]
table {%
8.75 0.613404088050314
9.25 0.613404088050314
};
\addplot [matplotcyan]
table {%
9.75 0.683372641509434
10.25 0.683372641509434
};
\addplot [matplotcyan]
table {%
10.75 0.667551100628931
11.25 0.667551100628931
};
\addplot [matplotcyan]
table {%
11.75 0.653596698113208
12.25 0.653596698113208
};
\addplot [matplotcyan]
table {%
12.75 0.683863993710692
13.25 0.683863993710692
};
\addplot [matplotcyan]
table {%
13.75 0.654775943396226
14.25 0.654775943396226
};
\addplot [matplotcyan]
table {%
14.75 0.686124213836478
15.25 0.686124213836478
};
\addplot [matplotcyan]
table {%
15.75 0.686615566037736
16.25 0.686615566037736
};
\addplot [matplotcyan]
table {%
16.75 0.750982704402516
17.25 0.750982704402516
};
\end{axis}

\end{tikzpicture}

%% file: figure_scripts/agnews_ramia_vs_mia_pii.tex
\begin{tikzpicture}
	\begin{axis}[
		xlabel={(Range) FPR}, ylabel={(Range) TPR},
		ylabel style={yshift=-0.3cm},
		xmin = 0, xmax = 1,
        ymin = 0, ymax = 1, yscale=1,
		xtick={0.0,0.2,0.4,0.6,0.8,1.0}, xticklabels={0.0,0.2,0.4,0.6,0.8,1.0},
		grid = major, title style={yshift=0.9cm},
		legend style={at={(1.0, 0.0)},anchor=south east, font=\tiny}
		]	

		\addplot[solid, very thick, matplotblue, no marks] table[skip first n=1,x index=0, y index=1, col sep=comma] {"data/agnews/pii_mia.csv"};
		\addplot[solid, very thick, matplotred, no marks] table[skip first n=1,x index=0, y index=1, col sep=comma] {"data/agnews/pii_rmia.csv"};
		
		\addplot[densely dotted, thick, black] expression {x};
		
		\legend{MIA (AUC=0.579), RaMIA (AUC=0.590)}
	\end{axis}
\end{tikzpicture}

%% file: sections/sec_conclusion.tex
\section{Conclusion}
We have shown that traditional membership inference attacks (MIAs), which only consider information leakage at exact training points, fail to capture the broader privacy risks in similar but not identical data. We show that when the query shifts from exact training points to nearby points, MIA performance degrades drastically under the correct privacy notion. To address this shortcoming, we introduced range membership inference attacks that evaluates whether a given range contains any training data. Our formulation casts the problem as a composite hypothesis test and proposes a robust trimmed averaging approach to aggregate membership scores over a set of sampled points. 

Our proof-of-concept experiments on tabular, image and text datasets demonstrate that RaMIA outperforms  MIAs in realistic scenarios with non-exact queries. We acknowledge that RaMIA presents a new technical challenge and that our current formulation and solution are not optimal; rather, they are intended as a starting point to stimulate further research in this promising and meaningful direction. While there is ample room for improving the sampling process and refining the range function design, our work highlights the potential of range-based privacy auditing and motivates future efforts to develop more powerful and robust RaMIA strategies.

We hope that our work can encourage privacy researchers and practitioners to re-examine the conventional MIA paradigm and consider range-based approaches as a more realistic and comprehensive tool for privacy auditing.

%% file: sections/acknowledgement.tex
\section*{Acknowledgement}
The authors would like to thank Milad Nasr for the discussion and exploration at the early stage of this project. This research is supported by the Ministry of Education, Singapore, Academic Research Fund (AcRF) Tier 1 (A-8001610-00-00) and Tier 2 (MOE-T2EP20223-0015).

%% file: sections/app_attacks.tex
\subsection{Attack algorithms} \label{app:attacks}
In this section, we explain the details of the membership inference attack algorithms used in our experiments. 
\paragraph{LOSS} LOSS \cite{yeom2018privacy} computes loss values as a proxy of membership score on given points: $\text{MIA}(x;\theta)=\mathit{l}(x;\theta)$. To compute the likelihood, an easy way is to take the exponential of the negative of the loss $\probP=\exp^{-\mathit{l}}$.
\paragraph{RMIA} RMIA \cite{zarifzadeh2023low} computes membership score by applying chain rule: $\probP{(\theta|x)} = \frac{\probP{(x|\theta)} \probP{(\theta)}}{\probP{(x)}}$. The score is then compared with all available population data points to obtain the percentage of population points being dominated by the given point: $\probP_{z \in Z}{[\frac {\probP{(\theta|x)}}{\probP{(\theta|z)}} \geq \gamma]}$, where the term $\probP{(\theta)}$ will cancel out with each other. The normalizing constant $\probP{P(x)}$ is computed with reference models: $\probP{(x)}=0.5\mathbb{E}_{\theta_\text{IN}} \probP{(x|\theta_\text{IN})} + 0.5\mathbb{E}_{\theta_\text{OUT}} \probP{(x|\theta_\text{OUT})}$. In its offline version, the in models are unavailable. In this case, the former probabilities are approximated by the latter term $\probP_\text{IN} = a \probP_\text{OUT} + (1-a)$. The hyperparameter $\alpha$ is chosen based on the reference models. Specifically, one reference model is chosen as the temporary target model, and the rest are used to attack it. The value of $\alpha$ is chosen to be the best-performing value under this setting, obtained via a simple sweeping. In our experiment, we use the offline attack only. The $\alpha$ values for Purchase-100 and CIFAR-10 are taken from \cite{zarifzadeh2023low}. For CelebA, we set it to be 0.33. For AG News, we set it to be 1.0.
\paragraph{LiRA} LiRA \cite{carlini2022membership} constructis IN and OUT distribution of model outputs for each query point. Then the membership score is defined to be the ratio between the pdf values under the IN and OUT distributions: $\text{MIA}(x;\theta)=\frac{\probP{(\theta(x) | \theta_\text{IN}(x))}}{\probP{(\theta(x) | \theta_\text{OUT}(x))}}$. 

%% file: sections/app_more_results.tex
\subsection{RaMIA with LiRA}
In this section, we present more results. Firstly, we present the improvement of RaMIA over MIA using LiRA as the membership testing algorithm in Fig \ref{fig:attack_performance_lira}.

\begin{figure*}[h]
	\centering
	\begin{subfigure}[b]{0.49\textwidth}
		\centering
		\if\compileFigures1
		\input{figure_scripts/purchase_ramia_vs_mia_clean_lira.tex}
		\else
		\includegraphics[]{fig/\filename-figure\thefigureNumber.pdf}
		\stepcounter{figureNumber}
		\fi
		\caption{Purchase-100 (m=10)}
		\label{fig:attack_purchase_lira}
	\end{subfigure}
	\hfill
	\begin{subfigure}[b]{0.49\textwidth}
		\centering
		\if\compileFigures1
		\input{figure_scripts/celeba_ramia_vs_mia_clean_lira.tex}
		\else
		\includegraphics[]{fig/\filename-figure\thefigureNumber.pdf}
		\stepcounter{figureNumber}
		\fi
		\caption{CelebA}
		\label{fig:attack_celeba_lira}
	\end{subfigure}
	\vskip\baselineskip
	\begin{subfigure}[b]{0.49\textwidth}
		\centering
		\if\compileFigures1
		\input{figure_scripts/cifar10_ramia_vs_mia_clean_lira.tex}
		\else
		\includegraphics[]{fig/\filename-figure\thefigureNumber.pdf}
		\stepcounter{figureNumber}
		\fi
		\caption{CIFAR-10}
		\label{fig:attack_cifar_lira}
	\end{subfigure}
	\hfill
	\begin{subfigure}[b]{0.49\textwidth}
		\centering
		\if\compileFigures1
		\input{figure_scripts/agnews_ramia_vs_mia_clean_lira.tex}
		\else
		\includegraphics[]{fig/\filename-figure\thefigureNumber.pdf}
		\stepcounter{figureNumber}
		\fi
		\caption{AG News (d=5)}
		\label{fig:attack_agnews_lira}
	\end{subfigure}
	\caption{RaMIAs perform better than MIAs on points that are close to original points but not exactly the same, using LiRA as the membership testing algorithm.}
	\label{fig:attack_performance_lira}
\end{figure*}

\subsection{Trimmed Means vs GLRT and Bayes Factor}
We explained that we modified the established statistical methods, GLRT and Bayes Factor, to solve our composite hypothesis testing problem due to the unreliability of MIA algorithms and the presence of noise. In this section, we show the comparison between our trimmed means and GLRT and Bayes Factor in Fig.

\begin{figure*}[h]
	\centering
	\begin{subfigure}[b]{0.49\textwidth}
		\centering
		\if\compileFigures1
		\input{figure_scripts/purchase_glrt_bayes.tex}
		\else
		\includegraphics[]{fig/\filename-figure\thefigureNumber.pdf}
		\stepcounter{figureNumber}
		\fi
		\caption{Purchase-100 (m=10)}
		\label{fig:purchase_glrt_bayes}
	\end{subfigure}
	\hfill
	\begin{subfigure}[b]{0.49\textwidth}
		\centering
		\if\compileFigures1
		\input{figure_scripts/celeba_glrt_bayes.tex}
		\else
		\includegraphics[]{fig/\filename-figure\thefigureNumber.pdf}
		\stepcounter{figureNumber}
		\fi
		\caption{CelebA}
		\label{fig:celeba_glrt_bayes}
	\end{subfigure}
	\vskip\baselineskip
	\begin{subfigure}[b]{0.49\textwidth}
		\centering
		\if\compileFigures1
		\input{figure_scripts/cifar10_glrt_bayes.tex}
		\else
		\includegraphics[]{fig/\filename-figure\thefigureNumber.pdf}
		\stepcounter{figureNumber}
		\fi
		\caption{CIFAR-10}
		\label{fig:cifar_glrt_bayes}
	\end{subfigure}
	\hfill
	\begin{subfigure}[b]{0.49\textwidth}
		\centering
		\if\compileFigures1
		\input{figure_scripts/agnews_glrt_bayes.tex}
		\else
		\includegraphics[]{fig/\filename-figure\thefigureNumber.pdf}
		\stepcounter{figureNumber}
		\fi
		\caption{AG News (d=5)}
		\label{fig:agnews_glrt_bayes}
	\end{subfigure}
	\caption{Our trimmed means approach is better than GLRT and Bayes Factor. Note for \ref{fig:celeba_glrt_bayes}, the Bayes Factor approach is the common user-level inference aggregation method.}
	\label{fig:vs_glrt_bayes}
\end{figure*}

%% file: figure_scripts/purchase_ramia_vs_mia_clean_lira.tex
\begin{tikzpicture}
	\begin{axis}[
		scale=0.9,
		xlabel={(Range) FPR}, ylabel={(Range) TPR},
		ylabel style={yshift=-0.3cm},
		xmin = 0, xmax = 1,
        ymin = 0, ymax = 1, yscale=1,
		xtick={0.0,0.2,0.4,0.6,0.8,1.0}, xticklabels={0.0,0.2,0.4,0.6,0.8,1.0},
		grid = major, title style={yshift=0.9cm},
		legend style={at={(1.0, 0.0)},anchor=south east, font=\tiny}
		]

		\addplot[solid, thick, matplotblue, no marks] table[skip first n=1,x index=0, y index=1, col sep=comma] {"data/purchase/lira_ad_guess_r10.csv"};

		\addplot[solid, very thick, matplotred, no marks] table[skip first n=1,x index=0, y index=1, col sep=comma] {"data/purchase/range_lira_r10.csv"};
		
		\addplot[densely dotted, thick, black] expression {x};
		
		\legend{MIA (AUC=0.637), RaMIA (AUC=0.643)}
	\end{axis}
\end{tikzpicture}

%% file: figure_scripts/celeba_ramia_vs_mia_clean_lira.tex
\begin{tikzpicture}
	\begin{axis}[
		scale=0.9,
		xlabel={(Range) FPR}, ylabel={(Range) TPR},
		ylabel style={yshift=-0.3cm},
		xmin = 0, xmax = 1,
        ymin = 0, ymax = 1, yscale=1,
		xtick={0.0,0.2,0.4,0.6,0.8,1.0}, xticklabels={0.0,0.2,0.4,0.6,0.8,1.0},
		grid = major, title style={yshift=0.9cm},
		legend style={at={(1.0, 0.0)},anchor=south east, font=\tiny}
		]	



		\addplot[solid, very thick, matplotblue, no marks] table[skip first n=1,x index=0, y index=1, col sep=comma] {"data/celeba/lira_norange.csv"};
	\addplot[solid, very thick, matplotred, no marks] table[skip first n=1,x index=0, y index=1, col sep=comma] {"data/celeba/range_lira.csv"};
		
		\addplot[densely dotted, thick, black] expression {x};
		
		\legend{MIA (AUC=0.561), RaMIA (AUC=0.609)}
	\end{axis}
\end{tikzpicture}

%% file: figure_scripts/cifar10_ramia_vs_mia_clean_lira.tex
\begin{tikzpicture}
	\begin{axis}[
		scale=0.9,
		xlabel={(Range) FPR}, ylabel={(Range) TPR},
		ylabel style={yshift=-0.3cm},
		xmin = 0, xmax = 1,
        ymin = 0, ymax = 1, yscale=1,
		xtick={0.0,0.2,0.4,0.6,0.8,1.0}, xticklabels={0.0,0.2,0.4,0.6,0.8,1.0},
		grid = major, title style={yshift=0.9cm},
		legend style={at={(1.0, 0.0)},anchor=south east, font=\tiny}
		]	

		\addplot[solid, very thick, matplotblue, no marks] table[skip first n=1,x index=0, y index=1, col sep=comma] {"data/cifar/aug_lira_norange_wrn.csv"};
		\addplot[solid, very thick, matplotred, no marks] table[skip first n=1,x index=0, y index=1, col sep=comma] {"data/cifar/aug_lira_range_wrn.csv"};
		
		\addplot[densely dotted, thick, black] expression {x};
		
		\legend{MIA (AUC=0.617), RaMIA (AUC=0.628)}
	\end{axis}
\end{tikzpicture}

%% file: figure_scripts/agnews_ramia_vs_mia_clean_lira.tex
\begin{tikzpicture}
	\begin{axis}[
		scale=0.9,
		xlabel={(Range) FPR}, ylabel={(Range) TPR},
		ylabel style={yshift=-0.3cm},
		xmin = 0, xmax = 1,
        ymin = 0, ymax = 1, yscale=1,
		xtick={0.0,0.2,0.4,0.6,0.8,1.0}, xticklabels={0.0,0.2,0.4,0.6,0.8,1.0},
		grid = major, title style={yshift=0.9cm},
		legend style={at={(1.0, 0.0)},anchor=south east, font=\tiny}
		]	

		\addplot[solid, very thick, matplotblue, no marks] table[skip first n=1,x index=0, y index=1, col sep=comma] {"data/agnews/norange_lira_subset_a5.csv"};
	
		\addplot[solid, very thick, matplotred, no marks] table[skip first n=1,x index=0, y index=1, col sep=comma] {"data/agnews/range_lira_subset_a5.csv"};
		
		\addplot[densely dotted, thick, black] expression {x};
		
		\legend{MIA (AUC=0.519), RaMIA (AUC=0.557)}
	\end{axis}
\end{tikzpicture}

%% file: figure_scripts/purchase_glrt_bayes.tex
\begin{tikzpicture}
	\begin{axis}[
		scale=0.9,
		xlabel={(Range) FPR}, ylabel={(Range) TPR},
		ylabel style={yshift=-0.3cm},
		xmin = 0, xmax = 1,
        ymin = 0, ymax = 1, yscale=1,
		xtick={0.0,0.2,0.4,0.6,0.8,1.0}, xticklabels={0.0,0.2,0.4,0.6,0.8,1.0},
		grid = major, title style={yshift=0.9cm},
		legend style={at={(1.0, 0.0)},anchor=south east, font=\tiny}
		]

		\addplot[solid, very thick, matplotblue, no marks] table[skip first n=1,x index=0, y index=1, col sep=comma] {"data/purchase/range_rmia_bottom6_r10.csv"};

		\addplot[solid, very thick, matplotred, no marks] table[skip first n=1,x index=0, y index=1, col sep=comma] {"data/purchase/range_rmia_glrt_r10.csv"};
		
		\addplot[solid, very thick, matplotgreen, no marks] table[skip first n=1,x index=0, y index=1, col sep=comma] {"data/purchase/range_rmia_bayes_r10.csv"};
		
		\addplot[dotted, very thick, matplotblue, no marks] table[skip first n=1,x index=0, y index=1, col sep=comma] {"data/purchase/range_lira_r10.csv"};
		
		\addplot[dotted, very thick, matplotred, no marks] table[skip first n=1,x index=0, y index=1, col sep=comma] {"data/purchase/range_lira_r10_glrt.csv"};
		
		\addplot[dotted, very thick, matplotgreen, no marks] table[skip first n=1,x index=0, y index=1, col sep=comma] {"data/purchase/range_lira_r10_bayes.csv"};
		
		\addplot[densely dotted, thick, black] expression {x};
		
		\legend{RMIA/Trimmed (AUC=0.657), RMIA/GLRT (AUC=0.657), RMIA/Bayes (AUC=0.652), LiRA/Trimmed(AUC=0.643), LiRA/GLRT (AUC=0.630), LiRA/Bayes (AUC=0.634)}
	\end{axis}
\end{tikzpicture}

%% file: figure_scripts/celeba_glrt_bayes.tex
\begin{tikzpicture}
	\begin{axis}[
		scale=0.9,
		xlabel={(Range) FPR}, ylabel={(Range) TPR},
		ylabel style={yshift=-0.3cm},
		xmin = 0, xmax = 1,
		ymin = 0, ymax = 1, yscale=1,
		xtick={0.0,0.2,0.4,0.6,0.8,1.0}, xticklabels={0.0,0.2,0.4,0.6,0.8,1.0},
		grid = major, title style={yshift=0.9cm},
		legend style={at={(1.0, 0.0)},anchor=south east, font=\tiny}
		]

		\addplot[solid, very thick, matplotblue, no marks] table[skip first n=1,x index=0, y index=1, col sep=comma] {"data/celeba/range_rmia.csv"};
		
		\addplot[solid, very thick, matplotred, no marks] table[skip first n=1,x index=0, y index=1, col sep=comma] {"data/celeba/range_rmia_glrt.csv"};
		
		\addplot[solid, very thick, matplotgreen, no marks] table[skip first n=1,x index=0, y index=1, col sep=comma] {"data/celeba/range_rmia_bayes.csv"};
		
		\addplot[dotted, very thick, matplotblue, no marks] table[skip first n=1,x index=0, y index=1, col sep=comma] {"data/celeba/range_lira.csv"};
		
		\addplot[dotted, very thick, matplotred, no marks] table[skip first n=1,x index=0, y index=1, col sep=comma] {"data/celeba/range_lira_glrt.csv"};
		
		\addplot[dotted, very thick, matplotgreen, no marks] table[skip first n=1,x index=0, y index=1, col sep=comma] {"data/celeba/range_lira_bayes.csv"};
		
		\addplot[densely dotted, thick, black] expression {x};
		
		\legend{RMIA/Trimmed (AUC=0.605), RMIA/GLRT (AUC=0.587), RMIA/Bayes (AUC=0.525), LiRA/Trimmed(AUC=0.601), LiRA/GLRT (AUC=0.585), LiRA/Bayes (AUC=0.523)}
	\end{axis}
\end{tikzpicture}

%% file: figure_scripts/cifar10_glrt_bayes.tex
\begin{tikzpicture}
	\begin{axis}[
		scale=0.9,
		xlabel={(Range) FPR}, ylabel={(Range) TPR},
		ylabel style={yshift=-0.3cm},
		xmin = 0, xmax = 1,
		ymin = 0, ymax = 1, yscale=1,
		xtick={0.0,0.2,0.4,0.6,0.8,1.0}, xticklabels={0.0,0.2,0.4,0.6,0.8,1.0},
		grid = major, title style={yshift=0.9cm},
		legend style={at={(1.0, 0.0)},anchor=south east, font=\tiny}
		]

		\addplot[solid, very thick, matplotblue, no marks] table[skip first n=1,x index=0, y index=1, col sep=comma] {"data/cifar/aug_rmia_range_wrn.csv"};
		
		\addplot[solid, very thick, matplotred, no marks] table[skip first n=1,x index=0, y index=1, col sep=comma] {"data/cifar/aug_rmia_range_wrn_glrt.csv"};
		
		\addplot[solid, very thick, matplotgreen, no marks] table[skip first n=1,x index=0, y index=1, col sep=comma] {"data/cifar/aug_rmia_range_wrn_bayes.csv"};
		
		\addplot[dotted, very thick, matplotblue, no marks] table[skip first n=1,x index=0, y index=1, col sep=comma] {"data/cifar/aug_lira_range_wrn.csv"};
		
		\addplot[dotted, very thick, matplotred, no marks] table[skip first n=1,x index=0, y index=1, col sep=comma] {"data/cifar/aug_lira_range_wrn_glrt.csv"};
		
		\addplot[dotted, very thick, matplotgreen, no marks] table[skip first n=1,x index=0, y index=1, col sep=comma] {"data/cifar/aug_lira_range_wrn_bayes.csv"};
		
		\addplot[densely dotted, thick, black] expression {x};
		
		\legend{RMIA/Trimmed (AUC=0.621), RMIA/GLRT (AUC=0.528), RMIA/Bayes (AUC=0.614), LiRA/Trimmed(AUC=0.628), LiRA/GLRT (AUC=0.609), LiRA/Bayes (AUC=0.564)}
	\end{axis}
\end{tikzpicture}

%% file: figure_scripts/agnews_glrt_bayes.tex
\begin{tikzpicture}
	\begin{axis}[
		scale=0.9,
		xlabel={(Range) FPR}, ylabel={(Range) TPR},
		ylabel style={yshift=-0.3cm},
		xmin = 0, xmax = 1,
		ymin = 0, ymax = 1, yscale=1,
		xtick={0.0,0.2,0.4,0.6,0.8,1.0}, xticklabels={0.0,0.2,0.4,0.6,0.8,1.0},
		grid = major, title style={yshift=0.9cm},
		legend style={at={(1.0, 0.0)},anchor=south east, font=\tiny}
		]

		\addplot[solid, very thick, matplotblue, no marks] table[skip first n=1,x index=0, y index=1, col sep=comma] {"data/agnews/range_rmia_subset_a5.csv"};
		
		\addplot[solid, very thick, matplotred, no marks] table[skip first n=1,x index=0, y index=1, col sep=comma] {"data/agnews/range_rmia_subset_a5_glrt.csv"};
		
		\addplot[solid, very thick, matplotgreen, no marks] table[skip first n=1,x index=0, y index=1, col sep=comma] {"data/agnews/range_rmia_subset_a5_bayes.csv"};
		
		\addplot[dotted, very thick, matplotblue, no marks] table[skip first n=1,x index=0, y index=1, col sep=comma] {"data/agnews/range_lira_subset_a5.csv"};
		
		\addplot[dotted, very thick, matplotred, no marks] table[skip first n=1,x index=0, y index=1, col sep=comma] {"data/agnews/range_lira_subset_a5_glrt.csv"};
		
		\addplot[dotted, very thick, matplotgreen, no marks] table[skip first n=1,x index=0, y index=1, col sep=comma] {"data/agnews/range_lira_subset_a5_bayes.csv"};
		
		\addplot[densely dotted, thick, black] expression {x};
		
		\legend{RMIA/Trimmed (AUC=0.599), RMIA/GLRT (AUC=0.587), RMIA/Bayes (AUC=0.568), LiRA/Trimmed(AUC=0.557), LiRA/GLRT (AUC=0.512), LiRA/Bayes (AUC=0.548)}
	\end{axis}
\end{tikzpicture}